\documentclass[11pt]{article}

\usepackage[preprint]{acl}

\usepackage{times}
\usepackage{latexsym}

\usepackage{booktabs}
\usepackage{amsmath}
\usepackage{multirow}
\usepackage{subcaption}
\usepackage{xcolor}
\newcommand{\dauc}[1]{%
  \ifdim#1pt>0pt \textcolor{teal}{$+#1$}%
  \else\ifdim#1pt<0pt \textcolor{red}{$#1$}%
  \else $\pm0.000$\fi\fi}
\newcommand{\deer}[1]{%
  \ifdim#1pt>0pt \textcolor{red}{$+#1$}%
  \else\ifdim#1pt<0pt \textcolor{teal}{$#1$}%
  \else $\pm0.000$\fi\fi}

\usepackage[T1]{fontenc}

\usepackage[utf8]{inputenc}

\usepackage{microtype}

\usepackage{inconsolata}

\usepackage{graphicx}

\title{SPARC: Spoken Profile Authorship Recognition Corpus}

\author{Efthymios Tsaprazlis$^1$\, Bowen Yi$^1$ \, Tiantian Feng$^1$ \\
\textbf{Junghwan Kim$^2$ \, David Jurgens$^2$ \, Shrikanth Narayanan$^1$}  \\
University of Southern California$^1$ \, University of Michigan$^2$ \\
\texttt{tsaprazl@usc.edu}
}

\begin{document}
\maketitle
\begin{abstract}

Speech anonymization has focused on the acoustic signal, leaving open how much speaker information the transcribed words themselves retain. In this work, we study pairwise speaker and demographic-attribute verification from transcribed speech. We introduce \textbf{SPARC} (Spoken Profile Authorship Recognition Corpus), a benchmark built from two podcast corpora comprising over 12{,}500 speakers, 1.5M quality-filtered utterances, and speaker-level inferred or source-provided labels for sex, age, and accent, and evaluate five attribution models and three LLMs on identity- and attribute-verification trials with episode- and metadata-based topic proxies. LLM adversaries distinguish same-speaker pairs from text alone. Attribute-label agreement is harder to verify, but LLMs perform sex- and accent-label verification above chance. Finally, releasing only short probe texts (50 words) degrades the strongest adversaries the most, offering a partial reduction in observed linkage.\footnote{Our code and data are available at \url{https://huggingface.co/datasets/timtsapras23/SPARC} and \url{https://github.com/bowenyi-pierre/SPARC}.}

\end{abstract}

\section{Introduction}

Conversational language is one of the most abundant forms of personal data. People speak to voice assistants, dictate messages, attend recorded meetings, and publish podcasts; even much social-media text approaches the register of speech. Regulatory frameworks have responded to the acoustic side of this reality: under the GDPR \cite{regulation2016regulation}, voice recordings constitute biometric personal data, and voiceprint-based speaker recognition is treated as a well-understood re-identification risk~\citep{nautsch2019gdpr,tigunova-etal-2020-reddust}. Far less attention has been paid to what lies beyond acoustic signals, to the words and phrases themselves, and how language is being used.

Transcription is often understood as separating \emph{what} was said from \emph{how} it sounded, and this distinction carries privacy weight: once pitch, timbre, and other acoustic characteristics are removed, conventional speaker-recognition systems no longer apply. Yet a transcript preserves word choice, syntax, discourse markers, and conversational habits, and these linguistic traces may suffice to link an anonymous transcript to a known speaker or to infer sensitive characteristics of that speaker~\citep{aggazzotti-etal-2024-authorship}. Prior work shows that speakers can be distinguished in human-transcribed conversations and that attribution survives ASR errors~\citep{aggazzotti-etal-2024-authorship,aggazzotti-etal-2025-impact}, but reports that performance declines under conversational topic controls. Whether this pattern holds against modern LLM adversaries, and whether \emph{demographic attribute labels} can be verified alongside identity, remain open questions.

To answer them, we introduce \textbf{SPARC} (Spoken Profile Authorship Recognition Corpus), a transcript-privacy benchmark built from two large-scale podcast corpora comprising over 12{,}500 speakers, 1.5M quality-filtered utterances, and speaker-level inferred or source-provided labels for sex, age, and accent. SPARC pairs speaker- and attribute-verification trials under a unified protocol, using episode- and metadata-based topic proxies to assess speaker-linked information in transcripts.

Our comprehensive benchmark results show that transcribed speech
retains speaker-linked information.
Overall, embedding-based attackers degrade in the within-episode condition (0.350 to 0.460 EER, approaching the 0.5 chance rate), but LLM adversaries retain speaker-verification signal in the same setting (0.322--0.345 EER) and reach 0.253--0.285 EER in the cross-episode condition on both corpora. Moreover, LLMs detect agreement on inferred sex and accent labels more reliably than the embedding models, while age-label agreement stays near chance. We further trace how leakage scales with the adversary's resources. Shortening the exposed probe text degrades the strongest attackers the most. Even an adversary with extensive reference material is bottlenecked by the probe. Limiting the amount of released text is thus a partial mitigation, although it attenuates rather than eliminates leakage. Finally, we show that average-case metrics could misstate the threat. Specifically, at a 1\% false-positive rate, two LLMs with nearly identical AUC and EER differ by more than $2\times$ in the fraction of positive trials they accept, so the models that matter most for privacy are not captured by AUC or EER alone.

Our contributions are: (i)~SPARC, a large-scale benchmark dataset for pairwise speaker and attribute-label verification in transcribed natural speech, with standardized trials across probe-length and enrollment budgets; (ii)~a systematic evaluation of embedding-based and LLM adversaries showing that LLMs retain speaker-verification signal under episode-context controls; (iii)~evidence that attribute-label agreement remains difficult to verify, though LLMs detect a coarse sex- and accent-label signal; (iv)~ablations showing that limiting exposed text degrades LLM attackers even under extensive enrollment, offering a practical but partial mitigation; and (v)~a low-FPR analysis demonstrating that average-case metrics mischaracterize trial-level linkage risk, with implications for how transcript privacy should be evaluated.

\section{Related Work}
\subsection{Text-Based Authorship Attribution and Verification}

Recent authorship attribution and verification systems learn author representations that transfer across documents and domains. LUAR, for instance, learns general-purpose author embeddings for open-set verification \citep{rivera-soto-etal-2021-learning}. A persistent concern is that such representations may encode what an author discusses rather than how they use language, motivating content-independent style representations, topic-regularized training, and probing studies that separate stylistic from topical information \citep{wegmann-etal-2022-author,sawatphol-etal-2022-topic,wang-etal-2023-authorship}. Most recently, LLMs have been applied to zero-shot, end-to-end verification and, beyond pairwise verification, to identity linking, extracting identity-relevant cues from unstructured profiles to retrieve candidate users across platforms \citep{huang-etal-2024-large,lermen2026usenix}.

A parallel line of work provides datasets and benchmarks for authorship analysis. VALLA standardizes datasets, splits, and metrics to compare attribution and verification methods \citep{tyo-etal-2023-valla}, and RAVEN tests whether verification models exploit topic-dependent shortcuts \citep{sawatphol-etal-2024-addressing}. Closer to our setting, HANSEN aggregates 17 human speech-transcript datasets, along with AI-generated spoken text, for attribution, verification, and human--AI detection \citep{tripto-etal-2023-hansen}. Most directly, \citet{aggazzotti-etal-2024-authorship} construct speaker-verification trials from human-transcribed conversations and find that performance declines once conversational topic is controlled; follow-up work examines how ASR errors affect transcript-based attribution \citep{aggazzotti-etal-2025-impact}. Text-only verification on VoicePrivacy data also reveals speaker-linked cues, partly through semantic keyword overlap~\citep{gaznepoglu25_interspeech}. SPARC complements these resources with large-scale, naturally occurring podcast transcripts, episode- and metadata-based controls, and a single protocol to evaluate speaker verification and attribute-label agreement.

\subsection{Personal Attribute Inference from Text}
Early efforts in author profiling focused on developing feature-based and neural classifiers for age and gender prediction, such as in the PAN author-profiling shared tasks \citep{rangel2013overview}.
Recent work reframes this capability as a privacy risk: \citet{staab2024beyond} show on PersonalReddit, a manually annotated dataset of 520 real Reddit profiles with eight attributes, that pretrained LLMs can infer personal information from text without relying on verbatim training-data memorization. Because the underlying profiles are sensitive, the dataset is not publicly released, making dataset construction a central challenge in this area. RedDust labels more than 300{,}000 Reddit users with profession, hobby, family status, age, and gender using high-precision self-disclosure patterns \citep{tigunova-etal-2020-reddust}, while SynthPAI uses LLM agents seeded with synthetic profiles to generate over 7{,}800 Reddit-like comments annotated with personal attributes, yielding a public, privacy-preserving benchmark whose conclusions are validated against real data \citep{yukhymenko2024synthetic}. These resources offer either broad scale or wide attribute inventories, but all concern authored social-media text, leaving attribute leakage in transcribed speech unexplored.

\subsection{Speech and Text Anonymization}
VoicePrivacy studies acoustic anonymization while preserving linguistic content~\citep{tomashenko2024voiceprivacy}. Text-anonymization work evaluates authorship obfuscation and privacy--utility trade-offs~\citep{loiseau-etal-2025-tarot,loiseau-etal-2025-tau,loiseau-etal-2026-adaptive, aggazzotti2026content}. SPARC evaluates existing transcripts rather than transforming them.

\section{Dataset}

\subsection{Source Corpora}
The SPARC dataset is drawn from two publicly available large-scale English podcast datasets.

\paragraph{NaturalVoices~\citep{du2025naturalvoices}} is a dataset of 5,049 hours of spontaneous podcast speech sourced from MSP-Podcast recordings, including meta-information at the utterance level with speaker identity, sex, speech quality, and transcripts. In total, the dataset has over 2,670 speakers across 6,790 episodes. We use only single-speaker speech segments from episodes released with a CC-BY 4.0 license, and we adopt the age and sex labels provided in the original metadata.

\paragraph{SPoRC~\citep{litterer2025mapping}} consists of over 1.1 million English podcast episodes collected from public RSS feeds between May and June 2020. Furthermore, it has a machine-diarized subset of over 220,000 hours that identifies 386,000 hosts and 535,000 guests. Because SPoRC does not provide utterance-level speaker metadata, we use its provided single-speaker episode subset. To control speech volume variation, we retain only the middle 50\% of episodes by transcript word count. We derive all demographic labels through automatic inference using speech data (Section \ref{sec:att_inf}). Together, NaturalVoices offers richer utterance-level annotations, and the retained SPoRC subset broadens speaker and episode coverage.

\subsection{Spoken Language Data Filtering}
\label{filtering}
We apply a unified multi-stage data cleaning pipeline to both corpora. First, utterances with an SNR $\geq$ 15 dB are kept for the remaining processing to ensure the quality of transcription. Specifically, we estimated SNR via WADA-SNR \cite{kim2008robust} for NaturalVoices and Brouhaha \cite{lavechin2023brouhaha, Bredin2020} for SPoRC. Second, podcast recordings often contain background music, which could confound our subsequent analyses. Therefore, segments with background music are removed using an Audio Spectrogram Transformer (AST) classifier \citep{gong2021ast}, retaining only utterances whose summed probability over 97 music-related categories from the AudioSet ontology \citep{gemmeke2017audio} is $\leq$ 0.1. Third, we use agreement between two state-of-the-art ASR models, Whisper-large-v3 \cite{radford2022whisper} and nvidia/canary-qwen-2.5b \cite{puvvada2024less}, as a reference-free proxy for transcription reliability. Notably, these two models achieved top-ranked positions on the Open ASR Leaderboard \cite{srivastav2025openasrleaderboardreproducible}. Let $w(a, b)$ denote the word error rate of hypothesis $a$ against reference $b$; we retain a segment only if $\max(w(a, b), w(b, a)) \leq 0.1$, using the symmetric maximum to penalize disagreement in either direction.
Transcripts from the Canary model are used as the final transcript for SPARC.

\subsection{Speaker Attribute Inference}
\label{sec:att_inf}
For NaturalVoices, we take the age and sex labels from the dataset, which are inferred by a pretrained model~\cite{burkhardt2023speech}. For SPoRC, we infer age and sex using Vox-Profile~\cite{feng2025vox}, a speech foundation-model benchmark that provides state-of-the-art classifiers for speaker traits. Accent labels in NaturalVoices and SPoRC are inferred with Voxlect~\cite{feng2025voxlect}, a benchmark for dialect classification including English. Utterance-level accent is predicted by a Whisper-large-v3 model over 16 English dialect classes. On the other hand, age and sex are predicted jointly by a WavLM-large model.

We aggregate to the speaker level by averaging the predicted probability vectors across all retained utterances from the same speaker and selecting the highest-scoring class for accent and sex; age is averaged directly. Dataset statistics of SPARC are reported in Table~\ref{tab:corpus_stats}. We find that there are slightly more male speakers in both sub-datasets. Moreover, most speakers are inferred to age between 30 to 60, whereas SPoRC contains a notably larger proportion of younger speakers (<30: 36.3\% vs.\ 15.7\%). On the other hand, North American accent accounts for roughly 77--78\% of speakers in both datasets, but the remaining speakers span 15 other English dialect categories with none exceeding 6\%. As a result, we binarize accent classes to North American vs.\ Other.

Three independent annotators judged 50 recordings per attribute, sampled to balance the attribute categories. Model--majority agreement was 98\% for sex, 88\% for accent, and 64\% for age (Appendix~\ref{app:human-validation}).

\begin{figure}[h]
    \centering
    \includegraphics[width=0.95\linewidth]{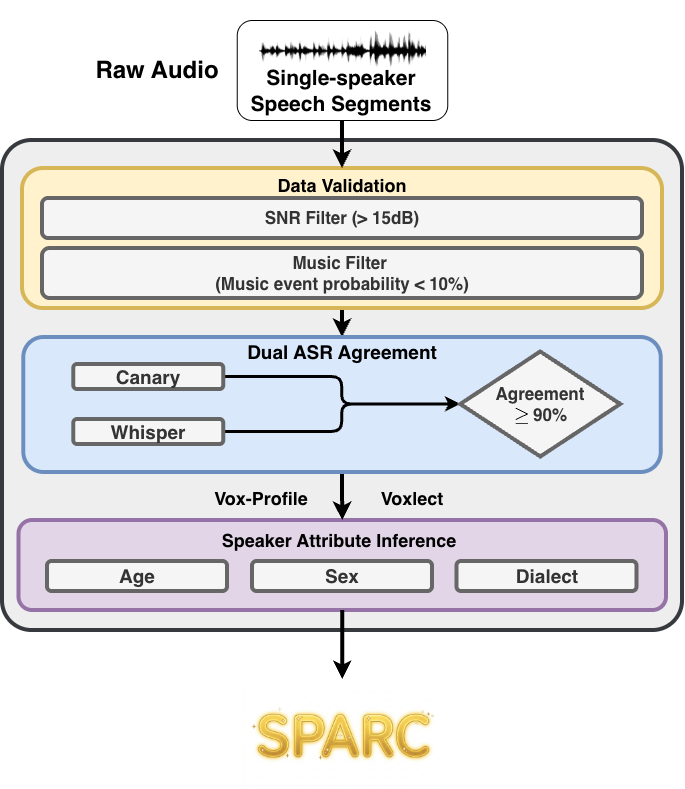}
    \vspace{-3mm}
    \caption{Overview of the SPARC construction pipeline. Eligible single-speaker speech from NaturalVoices and SPoRC is filtered for SNR and background music, retained when two ASR models agree by at least 90\%, and annotated with age, sex, and accent labels from source metadata or model predictions.}
    \label{fig:pipeline}
    \vspace{-3mm}
\end{figure}

\begin{table}[h]
\centering
\small
\begin{tabular}{lrr}
\toprule
& \textbf{NaturalVoices} & \textbf{SPoRC} \\
\midrule
Number of Speakers & 2,344 & 10,165 \\
Number of Episodes & 3,500 & 15,555 \\
Speaker--episode pairs & 6,766 & 15,555 \\
Number of Utterances & 779,494 & 795,591 \\
Avg.\ utt.\ / speaker & 332.6 & 78.3 \\
Avg.\ utt.\ / speaker--episode & 115.2 & 51.2 \\
Avg.\ age (yrs) & 41.5 & 37.9 \\
\midrule
\multicolumn{3}{l}{\textit{Sex (\%)}} \\
~~Male & 52.4 & 59.3 \\
~~Female & 47.6 & 40.7 \\
\midrule
\multicolumn{3}{l}{\textit{Age Group (\%)}} \\
~~<30 & 15.7 & 36.3 \\
~~30--60 & 76.2 & 53.1 \\
~~60+ & 8.1 & 10.6 \\
\midrule
\multicolumn{3}{l}{\textit{Accent (\%)}} \\
~~North American & 78.2 & 77.3 \\
~~Other & 21.8 & 22.7 \\
\bottomrule
\end{tabular}
\caption{Corpus statistics after filtering and attribute inference.
Sex and age are taken from dataset metadata for NaturalVoices and
inferred via Vox-Profile for SPoRC subset. Accent is binarized into
\textit{North American} vs.\ \textit{Other} due to the sparsity of remaining accent groups.}
\label{tab:corpus_stats}
\vspace{-3mm}
\end{table}

\section{Evaluation}

\subsection{Problem Setup}
\label{problem_setup}

We aim to investigate whether spoken language can provide information to characterize an individual’s identity or personal attributes. To this end, we formulate our problem similarly to established approaches in the Automatic Speaker Verification (ASV), adapting two evaluation tasks from VoxGuard \cite{11464033} (measuring \texttt{User Privacy} and \texttt{Attribute Privacy}). Concretely, we consider an adversarial setting in which the goal is to determine whether two sets of utterances originate from the same speaker, or whether they are produced by speakers sharing a specific inferred attribute label (e.g. sex). The adversary is provided with an \textit{enrollment} set, used to construct reference representations and a \textit{probe} utterance whose underlying identity or attribute membership is unknown. These are pairwise verification tasks, not identification from a population.

\subsection{Trial Generation}
\label{trial_gen}

We construct verification trials from NaturalVoices and SPoRC for four verification tasks: \textit{speaker identity}, \textit{sex}, \textit{age group}, and \textit{accent}. Throughout, a \emph{podcast series} contains multiple \emph{episodes}; we use \emph{podcast} to refer to a single episode. For each (speaker/attribute, podcast) pair, we form a probe by traversing the transcript from the beginning and accumulating complete utterances until its length falls within $\pm 5$ words of the target probe length $P$. For same-podcast trials, enrollment has the final $L$ non-probe utterances from the same podcast; for cross-podcast trials, it comprises the first $L$ utterances from another eligible podcast appearance. $L$ controls the amount of enrollment text. This construction defines four base conditions: \textit{Easy Positives} pair the same speaker within the same podcast, \textit{Hard Positives} pair the same speaker across podcasts, \textit{Easy Negatives} pair different speakers across podcasts, and \textit{Hard Negatives} pair different speakers within the same podcast. In NaturalVoices, the single-speaker constraint applies to segments rather than episodes: episodes may contain multiple speakers, enabling all four trial conditions. The retained SPoRC episodes contain only one speaker and therefore support only \textit{Hard Positives} and \textit{Easy Negatives}. For attribute tasks, positive and negative labels analogously indicate whether the two speakers share the evaluated attribute label. For reporting, we call within-episode trials \textit{same-topic} and cross-episode trials \textit{different-topic}. These are episode-based proxies because NaturalVoices lacks topic annotations or descriptions, and within-episode texts are expected to share more context. Comparing the conditions tests whether performance persists after controlling for episode co-membership.

This design serves two purposes. First, it allows us to measure model performance separately under easy and hard conditions and identify where models struggle. More importantly, it provides a diagnostic for spurious topical correlations: when probe and enrollment texts come from the same podcast, a model may perform well by exploiting shared content rather than speaker-specific linguistic style.
Across all SPoRC verification tasks, we control for topic by limiting episode-title/description similarity and matching positive and negative trials by this similarity and podcast category overlap. Speaker-verification negatives are additionally matched on sex, age group, and accent group. Full details and trial counts are provided in Appendix~\ref{app:trial-construction}.

\subsection{Model Scoring}
We evaluate eight models grouped by scoring approach. Five are representation-based baselines: four pretrained neural encoders---SentenceBERT~\citep{reimers2019sentence}, PCM~\citep{kim-etal-2025-leveraging}, CISR~\citep{wegmann-etal-2022-author}, and LUAR~\citep{rivera-soto-etal-2021-learning}---and a lexical TF--IDF~\citep{sparck1972statistical} baseline. These methods construct representations of the probe and enrollment texts and score each trial using cosine similarity. The remaining three models are prompted LLMs: GPT-5-mini, Llama-3.3-70B-Instruct, and Qwen3.5-9B. For each trial, we provide the probe and enrollment utterances and ask the model to return a score from 0 to 100 indicating the likelihood that the speakers share the same identity or evaluated attribute. Exact prompt templates and scoring details are provided in Appendix~\ref{app:inference_details}.

\subsection{Evaluation Metrics}
We report three complementary metrics that probe different regions of the ROC curve. Our primary metric is the \textbf{Equal Error Rate (EER)}: the error rate at the operating point where the false acceptance and false rejection rates coincide (lower is better; chance corresponds to 0.5). EER characterizes a concrete operating point rather than an average over thresholds, and it is not inflated by skewed positive-to-negative trial ratios, a property we rely on for the imbalanced accent trials. We additionally report \textbf{AUC}, the probability that a randomly chosen positive pair receives a higher score than a randomly chosen negative pair, as a complementary rank-based summary of overall separability. Because neither metric reflects the stringent operating points at which re-identification becomes consequential, we follow \citet{11464033} and report \textbf{TPR@FPR} at 1\% and 0.1\% false-positive rates, the fraction of positive trials accepted at thresholds with the stated false-positive rates.

\section{Results}

\begin{figure*}[t]
    \centering
    \includegraphics[width=0.98\linewidth]{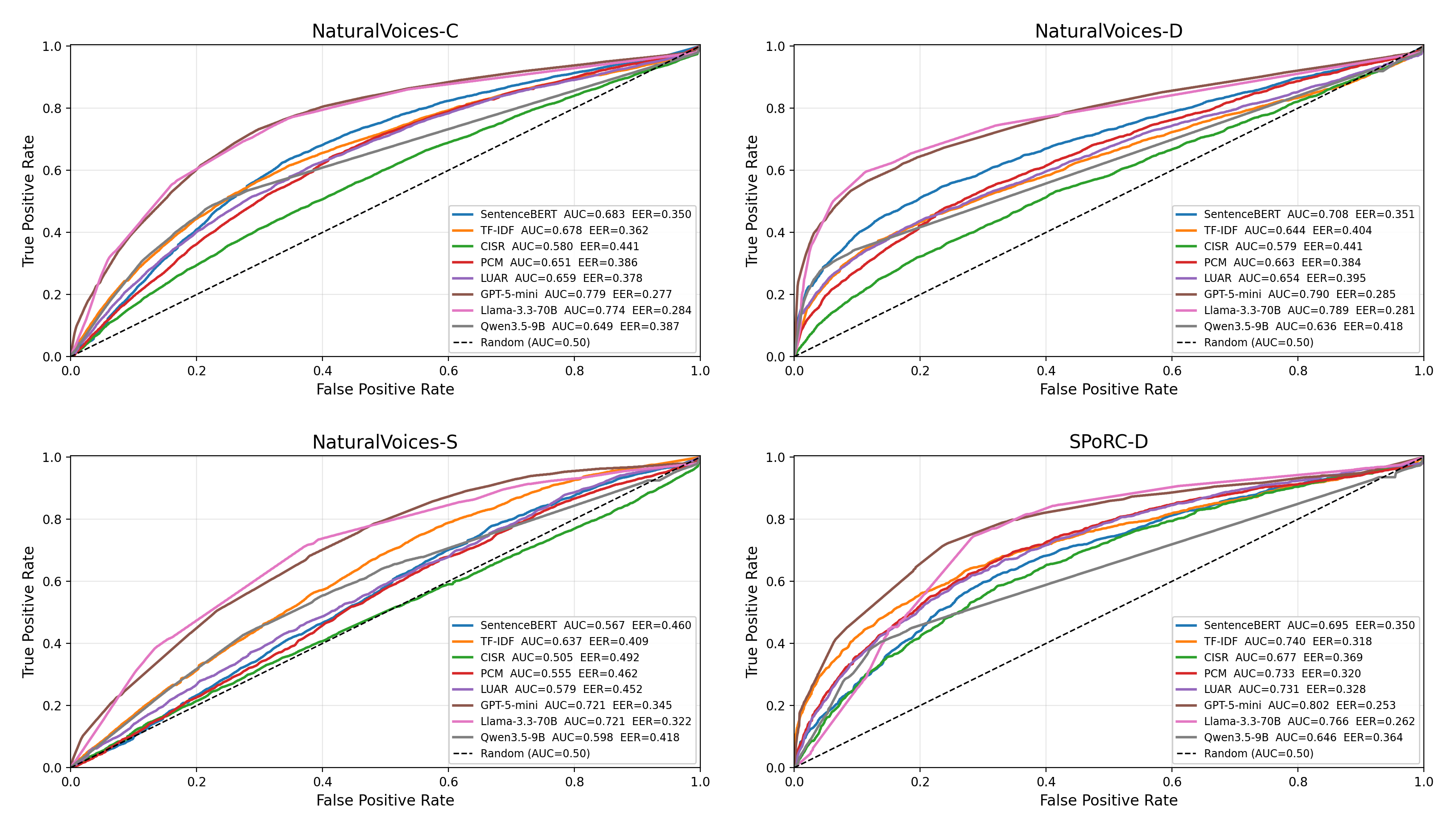}
    \vspace{-3mm}
    \caption{ROC curves for identity verification with 200-word probes and $L{=}20$ enrollment texts, for all eight models. NaturalVoices-C/S/D denote the combined, same-topic proxy (within-episode), and different-topic proxy (cross-episode) conditions and SPoRC-D is the different-topic condition of SPoRC.}
    \label{fig:plots}
    \vspace{-3mm}
\end{figure*}

\subsection{Identity Verification}

We evaluate identity verification in our base configuration, with approximately 200-word probes and ($L{=}20$) enrollment utterances. For clarity, the main-text discussion focuses on SentenceBERT as representative of the embedding-based attribution baselines and on GPT-5-mini and Llama-3.3-70B, the two strongest LLMs. Qwen3.5-9B performs comparably to the representation baselines (Figure~\ref{fig:plots}), while TF--IDF is discussed separately where it departs from their general pattern.
Figure~\ref{fig:plots} summarizes results across the three NaturalVoices conditions and the different-topic condition of SPoRC for all models.

Particularly, the different-topic condition most closely reflects real-world usage. In practice, samples of the same speaker will mostly come from different conversations and contexts, so verification must succeed across topical variation.
Overall, the finding in the different-topic condition is consistent across both corpora: LLMs including GPT-5-mini and Llama-3.3-70B attain EERs of 0.285 and 0.281 on NaturalVoices and 0.253 and 0.262 on SPoRC, far below the random chance baseline and well below SentenceBERT (0.351 and 0.350). The AUC results show the same ranking, with 0.77–0.80 for the LLMs versus 0.70 for SentenceBERT.

On the other hand, the same-topic condition serves as a diagnostic: episode co-membership is shared by positive and negative trials, though other contextual cues may remain. Here SentenceBERT degrades sharply, from 0.350 EER in the combined setting to 0.460, approaching chance, and the remaining embedding baselines show a similar collapse (LUAR: 0.378 to 0.452; PCM: 0.386 to 0.462; CISR to 0.492, effectively chance), consistent with reliance on episode context. This reflects a long-standing limitation of authorship attribution where models tend to conflate what is being said with who is saying it. The two strongest LLMs are far more robust. Llama-3.3-70B retains a 0.322 EER and GPT-5-mini 0.345 (both at 0.721 AUC). Still, the increase from the combined setting (0.284 and 0.277) shows that the LLMs are not fully immune to episode context. TF-IDF is a notable exception among the baselines where its EER rises only mildly in the within-episode condition (0.362 to 0.409), and it is the strongest non-LLM model on SPoRC (0.318 EER, 0.740 AUC), suggesting that surface word-choice statistics carry some speaker-linked signal, although it remains behind GPT-5-mini and Llama-3.3-70B in every condition.

\begin{table*}[t]
\centering
\small
\begin{tabular}{llcccccc}
\toprule
& & \multicolumn{2}{c}{\textbf{Combined}} & \multicolumn{2}{c}{\textbf{Same Topic}} & \multicolumn{2}{c}{\textbf{Diff.\ Topic}} \\
\cmidrule(lr){3-4}\cmidrule(lr){5-6}\cmidrule(lr){7-8}
\textbf{Attribute} & \textbf{Model} & \textbf{EER} $\downarrow$ & \textbf{AUC} $\uparrow$ & \textbf{EER} $\downarrow$ & \textbf{AUC} $\uparrow$ & \textbf{EER} $\downarrow$ & \textbf{AUC} $\uparrow$ \\
\midrule
\multirow{3}{*}{Biological Sex}
  & SentenceBERT  & 0.468 & 0.543 & 0.508 & 0.490 & 0.484 & 0.515 \\
  & GPT-5-mini    & \textbf{0.379} & \textbf{0.664} & 0.437 & 0.608 & \textbf{0.377} & \textbf{0.669} \\
  & Llama-3.3-70B & 0.403 & 0.632 & \textbf{0.422} & \textbf{0.611} & 0.420 & 0.612 \\
\midrule
\multirow{3}{*}{Age Group}
  & SentenceBERT  & 0.447 & \textbf{0.576} & \textbf{0.461} & \textbf{0.545} & 0.469 & 0.548 \\
  & GPT-5-mini    & \textbf{0.442} & 0.572 & 0.496 & 0.510 & \textbf{0.459} & \textbf{0.557} \\
  & Llama-3.3-70B & 0.546 & 0.433 & 0.542 & 0.474 & 0.481 & 0.535 \\
\midrule
\multirow{3}{*}{Accent}
  & SentenceBERT  & 0.441 & 0.578 & 0.483 & 0.534 & 0.494 & 0.510 \\
  & GPT-5-mini    & \textbf{0.279} & \textbf{0.763} & 0.350 & \textbf{0.701} & \textbf{0.268} & \textbf{0.771} \\
  & Llama-3.3-70B & 0.315 & 0.720 & \textbf{0.332} & 0.699 & 0.320 & 0.707 \\
\bottomrule
\end{tabular}
\caption{Attribute verification results on NaturalVoices with 200-word probes and $L{=}20$ enrollment texts. Same- and different-topic labels denote within- and cross-episode proxies, respectively. Best result per condition within each attribute in bold.}
\label{tab:attr_res}
\vspace{-3mm}
\end{table*}

\subsection{Attribute Verification}

We next ask whether transcripts support inferred demographic-label verification, evaluating sex, age, and accent under the same three conditions. As before, we focus on SentenceBERT as the representative embedding-based model and on GPT-5-mini and Llama-3.3-70B. We report results on NaturalVoices here; on the SPoRC attribute trials models show similar behavior, and that analysis, together with full results for all systems, is provided in Appendix~\ref{app:full-results}.

\subsubsection{Sex-Label Verification}
Verification on sex reveals a sharp divide between model families. Embedding-based models operate near chance across all conditions, with SentenceBERT EERs of 0.468--0.508, indicating little separation on this label-verification task.
Instead, GPT-5-mini and Llama-3.3-70B reach 0.377 and 0.420 EER in the realistic different-topic setting. Moreover, they retain this advantage under the same-topic control (0.437 and 0.422 EER). Their performance in the within-episode comparison suggests a modest label-agreement signal, although the responsible linguistic cues remain unclear.

\subsubsection{Age Group}

The benchmark results show that age-group agreement is the most difficult attribute task, where performance remains near chance in all three main conditions. The best EERs are only 0.447 in the different-topic and 0.440 in the same-topic condition, both from PCM (Table~\ref{tab:age_full}); among the main-text models, GPT-5-mini reaches 0.459 in the different-topic setting. Embedding-model AUCs in the different-topic setting range from 0.517 to 0.556, and Llama-3.3-70B falls below chance in the combined condition (0.546 EER, 0.433 AUC), suggesting a systematic bias in its age judgments rather than useful sensitivity. Given these small and inconsistent effects, we find no reliable evidence that the tested systems can verify age-group agreement from transcripts.

\subsubsection{Accent/Dialect}
Accent yields the clearest two-tier separation of all attributes. Embedding-based models sit at chance in the different-topic setting (0.494 EER for SentenceBERT; 0.494--0.534 AUC across the embedding baselines) and gain little in the within-episode condition. LLMs, by contrast, extract a markedly stronger signal. For example, GPT-5-mini achieves a 0.268 EER (0.771 AUC) in the different-topic setting and 0.350 under the same-topic control; Llama-3.3-70B reaches 0.320 and 0.332 EER, the best same-topic result. These results show agreement on accent labels, but do not identify which linguistic cues the LLMs use. Because the task uses an imbalanced North American vs.\ non--North-American distinction, these results do not establish performance on individual dialects but only on this binary coarse distinction.

\begin{table*}[t]
\centering
\small
\setlength{\tabcolsep}{5pt}
\begin{tabular}{l rr rr rr rr}
\toprule
& \multicolumn{2}{c}{\textbf{Main - 200w / }$L$\textbf{=20}}
  & \multicolumn{2}{c}{$\Delta$ \textbf{200w / }$L$\textbf{=5}}
  & \multicolumn{2}{c}{$\Delta$ \textbf{50w / }$L$\textbf{=20}}
  & \multicolumn{2}{c}{$\Delta$ \textbf{50w / }$L$\textbf{=5}} \\
\cmidrule(lr){2-3}\cmidrule(lr){4-5}\cmidrule(lr){6-7}\cmidrule(lr){8-9}
\textbf{Model}
  & EER\,$(\downarrow)$ & AUC\,$(\uparrow)$
  & EER\,$(\downarrow)$ & AUC\,$(\uparrow)$
  & EER\,$(\downarrow)$ & AUC\,$(\uparrow)$
  & EER\,$(\downarrow)$ & AUC\,$(\uparrow)$\\
\midrule
SentenceBERT  & 0.351 & 0.708
  & \deer{ 0.024} & \dauc{-0.025} & \deer{ 0.038} & \dauc{-0.047} & \deer{ 0.046} & \dauc{-0.059} \\
GPT-5-mini    & 0.285 & \textbf{0.790}
  & \deer{ 0.041} & \dauc{-0.049} & \deer{ 0.065} & \dauc{-0.077} & \deer{ 0.076} & \dauc{-0.089} \\
Llama-3.3-70B & \textbf{0.281} & 0.789
  & \deer{ 0.044} & \dauc{-0.045} & \deer{ 0.047} & \dauc{-0.053} & \deer{ 0.064} & \dauc{-0.075} \\
\bottomrule
\end{tabular}
\caption{Identity verification ablations on NaturalVoices (different-topic).
Main results use probe\,$\approx$\,200 words and $L{=}20$; $\Delta$ columns show
the change when reducing probe length and/or enrollment size $L$. Best EER and AUC in the main setting in \textbf{bold}.}
\label{tab:id_ablation}
\end{table*}

\begin{table*}[t]
\centering
\small
\begin{tabular}{lcccc}
\toprule
\textbf{Model} & \textbf{EER} $\downarrow$ & \textbf{AUC} $\uparrow$ & \textbf{TPR@1\%} & \textbf{TPR@0.1\%} \\
\midrule
SentenceBERT  & 0.351 & 0.708 & 0.143 & 0.056 \\
GPT-5-mini    & 0.285 & \textbf{0.790} & \textbf{0.277} & \textbf{0.121} \\
Llama-3.3-70B & \textbf{0.281} & 0.789 & 0.133 & 0.011 \\
\bottomrule
\end{tabular}
\caption{Low-FPR identity verification on NaturalVoices (different-topic, 200-word probes, $L{=}20$). TPR at FPR${=}1\%$ and $0.1\%$; bootstrap 95\% CIs are tight for TPR@1\% (e.g.\ GPT-5-mini $[0.265,0.293]$) and wider at $0.1\%$. Best per column in bold.}
\label{tab:low_fpr}

\end{table*}

\subsection{Sensitivity to Reduced Information}
\label{sec:ablations}

A privacy-relevant adversary rarely observes abundant text. We therefore ask how speaker-verification performance scales with the two quantities under an adversary's control: probe length (200 vs.\ 50 words) and enrollment size ($L{=}20$ vs.\ $L{=}5$). Table~\ref{tab:id_ablation} reports the effects on NaturalVoices. At a fixed probe length, the enrollment-size comparisons use the same speaker pairs and probes; only the number of enrollment utterances changes.

\paragraph{Probe length is the binding constraint for verification.}
Shortening the probe to 50 words degrades every model, with larger losses for the two strongest LLMs than for SentenceBERT. GPT-5-mini's EER rises by 0.065 in the different-topic setting (0.285 to 0.350), against 0.047 for Llama-3.3-70B and 0.038 for SentenceBERT.
Even though verification becomes more challenging with decreased probe size, the LLMs remain well below chance and below the embedding baselines throughout. Therefore, short probes attenuate rather than eliminate leakage with the use of LLMs. Moreover, this degradation pattern persists in the different-topic setting, where episode co-membership offers no direct shortcut, consistent with weaker textual signal as text is withheld. Full ablation results for all conditions are provided in the Appendix \ref{app:full-results}. Limiting probe length is thus a real but partial mitigation.

\paragraph{Enrollment size is a weaker bottleneck.}
Reducing enrollment to $L{=}5$ costs less than shortening the probe, though the asymmetry varies by model. It is pronounced for GPT-5-mini ($+0.041$ vs.\ $+0.065$ EER) and SentenceBERT ($+0.024$ vs.\ $+0.038$), and modest for Llama-3.3-70B ($+0.044$ vs.\ $+0.047$). The direction is nonetheless consistent, and the asymmetry follows from the task structure. Enrollment utterances are averaged into a single reference representation and saturate quickly, whereas the probe is the sole evidence about the unknown speaker. For the LLMs, the enrollment set functions as in-context evidence that characterizes the candidate speaker, so shrinking it does erode the model's basis for judging identity, but only mildly. Capable models, and the larger ones in particular, are efficient at extracting regularities from text, and a handful of utterances already supplies enough signal to keep performance close to the full $L{=}20$ setting. For an adversary, then, additional reference material yields diminishing returns, while probe text remains decisive.

\paragraph{Restricting both compounds only mildly.}
Cutting both to 50 words and $L{=}5$ yields the largest degradation (GPT-5-mini $+0.076$, Llama-3.3-70B $+0.064$ EER), and the LLMs lose more in absolute terms than the embedding models ($+0.046$). The joint effect is sub-additive, however it is governed by the probe bottleneck, with only a modest further cost from reduced enrollment. The privacy reading is twofold. Constraining the exposed text is an effective lever against the strongest adversaries, since it removes the most from the models that leak the most; yet even at this minimal setting the LLMs retain 0.35--0.36 different-topic EER, still well below both the 0.5 chance rate and the embedding baselines. Speaker-verification performance can be reduced, but not eliminated, by withholding text alone.

\subsection{Beyond Average-Case Metrics}
\label{sec:low_fpr}

Because EER and AUC obscure performance at low false-positive rates, we report TPR at 1\% and 0.1\% FPR for the NaturalVoices different-topic base setting ($P=200,L=20$; Table~\ref{tab:low_fpr}). As throughout, we focus on the realistic different-topic condition of the base configuration (200-word probes, $L{=}20$).

The low-FPR view overturns the average-case ranking. GPT-5-mini and Llama-3.3-70B achieve very close EER and AUC in the different-topic setting (0.285 vs.\ 0.281 EER; 0.790 vs.\ 0.789 AUC), yet they diverge sharply where it matters. At $1\%$ FPR, GPT-5-mini accepts more than twice the fraction of positive trials (0.277 vs 0.133 TPR) and at $0.1\%$ FPR an order of magnitude more (0.121 vs 0.011), with non-overlapping bootstrap confidence intervals. Llama-3.3-70B's TPR falls more sharply at stringent thresholds, although score granularity may affect this comparison. GPT-5-mini, by contrast, accepts 27.7\% of positive trials at 1\% trial-level FPR.

Two further observations sharpen the picture. First, the average-case ordering does not survive. SentenceBERT, despite a clearly worse EER (0.351 vs.\ 0.281), matches Llama-3.3-70B at $1\%$ FPR (0.143 vs 0.133), so conclusions drawn from average-case metrics alone would misrank low-FPR performance. Second, the gap is partly mechanistic. Prompted LLMs emit coarsely quantized scores, and a model whose confidences concentrate on a few round values yields a step-like ROC whose resolution at low FPR is limited by score granularity rather than by the underlying signal; GPT-5-mini's smoother, better-separated high-confidence tail may contribute to its stronger low-FPR performance.

\section{Conclusion}

We introduced SPARC, a large-scale benchmark for pairwise speaker and attribute verification from speech transcripts and evaluated five attribution models and three LLMs across four verification tasks. Our results show that speaker-linked information persists under the evaluated episode and metadata controls. Removing the acoustic signal therefore does not completely remove the speaker information and speech privacy evaluations should consider linguistic content as well as voice. Moreover, attribute agreement remains challenging to verify, with a modest signal for biological sex and accent labels, while age-group agreement is near chance for every system. Shorter transcripts reduce speaker-verification performance, especially for the strongest models, but do not eliminate it. Moreover, similar EER and AUC scores can conceal large differences in trial-level TPR at low false-positive rates.
We hope SPARC supports the development and evaluation of transcript-level privacy defenses.

\section*{Limitations}

\paragraph{Data Scope}
SPARC contains only English podcast speech and retains relatively clean, single-speaker segments through SNR, music, and dual-ASR-agreement filters. The findings may therefore not generalize to noisier or overlapping speech, other conversational domains, or multilingual and code-switched settings.

\paragraph{Attribute Labels and Accent Evaluation}
SPARC's sex, age, and accent labels are model-derived rather than independently verified demographics. The category-balanced, 50-recording audit measures perceptual consistency on a small selected sample, not corpus-wide accuracy; agreement is weakest for age. Attribute results therefore measure agreement with proxy labels. Approximately 77--78\% of speakers are assigned a North American accent, and no other individual dialect exceeds 6\%, motivating the North American/Other grouping. This imbalanced binary task cannot support conclusions about fine-grained accent verification, especially for lower-resourced dialects~\citep{pan-etal-2025-analyzing}. Negative pairs also constitute only 19\% of the combined NaturalVoices accent trials, reducing statistical precision. A more balanced corpus is needed for individual-dialect evaluation.

\paragraph{Model and Score Scope}
We evaluate three LLMs using a single zero-shot scoring prompt. Because of inference cost, we obtain one response per LLM-trial pair and therefore do not quantify variability across repeated generations. In addition, we notice that LLMs' uncalibrated 0--100 scores often cluster at a few rounded values, limiting ROC resolution at low FPR. The findings are therefore specific to the evaluated model--prompt configurations; calibrated or likelihood-based scoring could produce different rankings.

\paragraph{Linguistic Cue Attribution}
We do not attribute individual decisions to lexical, syntactic, discourse, stylistic, or ASR-derived cues. Such analysis is needed to separate speaker habits from residual context, particularly for the coarse accent-label result.

\paragraph{Scope of Topic Control}

For NaturalVoices, episode membership is only a proxy for topical overlap: one episode may span multiple topics, while different episodes may discuss similar subjects. The within- and cross-episode conditions therefore control episode context rather than semantic topic identity. For SPoRC, title/description filtering and category matching provide additional metadata-based proxies but cannot eliminate residual topical overlap. In the original SPoRC speaker protocol, 81.4\% of positive trials are within-series whereas all negatives cross series; the additional cross-series control reduces this imbalance and yields higher EERs. Neither analysis isolates speaker style from correlated textual cues, including personal facts and speaker-relevant ASR errors~\citep{aggazzotti-etal-2024-authorship}.

\section*{Ethical Considerations}
SPARC is intended to audit privacy leakage from speech transcripts, but the capabilities it evaluates are inherently dual use. The recordings were collected from publicly available podcast feeds; for NaturalVoices, we retain only episodes released under CC-BY 4.0. Nevertheless, public availability does not imply that speakers anticipated automated linkage or demographic profiling. We therefore position SPARC as a resource for privacy research and discourage its use in forensic, law-enforcement, employment, eligibility, or other high-stakes decisions.

Because SPoRC does not provide verified age, sex, or accent labels, we use Vox-Profile and Voxlect as reproducible proxies~\citep{feng2025vox,feng2025voxlect}. Vox-Profile's strongest models report 98.0\% accuracy for binary speaker-sex classification and 69.5\% for three-way age-group classification, while the Voxlect Whisper-Large model used in our study reports 83.0\% accuracy for English dialect classification. We reduce utterance-level noise by averaging predictions across each speaker's retained utterances and report a small perceptual audit in Appendix~\ref{app:human-validation}. These labels nevertheless remain model-inferred, and source-benchmark performance or agreement on a selected audit sample may not transfer fully to SPARC. The binary sex and coarse age and accent categories should therefore not be interpreted as verified demographic identities. We use them only for trial construction and aggregate analysis, not for downstream profiling or individual-level decisions. Any reuse should preserve their inferred provenance and comply with the source licenses, Voxlect's Responsible AI License, and applicable privacy requirements.

\bibliography{custom}

\appendix
\section{Appendix}
\label{sec:appendix}
\subsection{Verification Trial Construction}
\label{app:trial-construction}
This section provides corpus-specific details supplementing the general trial-generation procedure described in Section~\ref{trial_gen}.

\subsubsection{NaturalVoices} NaturalVoices episodes may contain multiple speakers and therefore support all four trial types: Easy Positives, Hard Positives, Easy Negatives, and Hard Negatives. Speaker trials are labeled according to whether the probe and enrollment texts belong to the same speaker. For the demographic tasks, trials are labeled according to whether the speakers share the evaluated sex, age-group, or accent label. For each eligible probe speaker-episode pair, enrollment partners requiring a different episode or speaker are randomly sampled from the corresponding eligible pool using a fixed seed; positive and negative trials are not matched one-to-one by episode or topic. Demographic trials are then retained according to attribute agreement without artificially balancing the classes, so their counts reflect the resulting attribute distributions. For the NaturalVoices speaker ablation, trials are fixed when varying $L$ at a fixed $P$: probe and speaker pairs are unchanged, and only the supplied enrollment utterances differ.

\subsubsection{SPoRC} The retained SPoRC episodes are single-speaker. Therefore, different-speaker, same-episode trials cannot be constructed, and all negative trials span different podcast series (characterized by different RSS feeds). Positive speaker trials pair the same speaker across distinct episodes, which may originate from the same or different podcast series. Negative speaker trials pair different speakers from different podcast series and require the speakers to share the same sex, age-group, and accent labels, reducing the possibility that broad demographic differences distinguish the trial classes.

We use two metadata-based topic proxies. First, we concatenate each episode's title and description and represent the resulting text using L2-normalized TF--IDF features over unigrams and bigrams. Candidate episode pairs with cosine similarity greater than 0.10 are excluded. Second, we represent each podcast series' non-empty category labels as a set and compute the Jaccard similarity between the two category sets. After the TF--IDF filter, candidate pairs are stratified jointly by TF--IDF similarity and category overlap. For each positive trial, we sample one eligible negative trial from the same stratum, producing balanced trial classes with closely matched distributions of the two metadata-topic measures. We additionally exclude pairs containing exact utterance overlap and screen positive speaker trials for explicit speaker-name leakage.

Tables~\ref{tab:naturalvoices-trials} and \ref{tab:sporc-trials} summarize the resulting evaluation protocols. Across the four probe and enrollment configurations, SPoRC contains 65,174 task-specific trials. NaturalVoices contains 124,356 trials. In the SPoRC speaker task, 6,993 of the 8,587 positive trials (81.4\%) pair different episodes from the same podcast series and 1,594 (18.6\%) span podcast series; all 8,587 negative trials span podcast series. Mean title--description cosine similarity is 0.027 for positive trials and 0.026 for negative trials, while mean category Jaccard similarity is approximately 0.89 for both classes.

For the additional speaker-verification control in Appendix~\ref{app:cross-series}, we restrict positive trials to pairs from different series, as the negative pairs already cross series. Thus, series co-membership is unavailable as a direct class cue in this analysis, although other contextual differences may remain.

\begin{table}[t]
\centering
\tiny
\setlength{\tabcolsep}{5pt}
\begin{tabular}{llrrrrr}
\toprule
\textbf{Probe / enrollment} & \textbf{Task} & \textbf{EP} & \textbf{HP} & \textbf{EN} & \textbf{HN} & \textbf{Total} \\
\midrule
\multirow{4}{*}{$P=50,\ L=5$}
 & Speaker & 3,748 & 3,046 & 4,015 & 2,789 & 13,598 \\
 & Sex  & 1,748 & 1,965 & 2,050 & 1,041 & 6,804 \\
 & Age     & 2,222 & 2,956 & 1,059 &   567 & 6,804 \\
 & Accent  & 2,445 & 3,055 &   960 &   344 & 6,804 \\
\midrule
\multirow{4}{*}{$P=50,\ L=20$}
 & Speaker & 3,748 & 3,046 & 4,015 & 2,789 & 13,598 \\
 & Sex  & 1,748 & 1,965 & 2,050 & 1,041 & 6,804 \\
 & Age     & 2,222 & 2,956 & 1,059 &   567 & 6,804 \\
 & Accent  & 2,445 & 3,055 &   960 &   344 & 6,804 \\
\midrule
\multirow{4}{*}{$P=200,\ L=5$}
 & Speaker & 2,987 & 2,513 & 3,382 & 2,285 & 11,167 \\
 & Sex  & 1,388 & 1,618 & 1,764 &   897 & 5,667 \\
 & Age     & 1,795 & 2,506 &   876 &   490 & 5,667 \\
 & Accent  & 2,010 & 2,592 &   790 &   275 & 5,667 \\
\midrule
\multirow{4}{*}{$P=200,\ L=20$}
 & Speaker & 2,987 & 2,513 & 3,382 & 2,285 & 11,167 \\
 & Sex  & 1,388 & 1,618 & 1,764 &   897 & 5,667 \\
 & Age     & 1,795 & 2,506 &   876 &   490 & 5,667 \\
 & Accent  & 2,010 & 2,592 &   790 &   275 & 5,667 \\
\bottomrule
\end{tabular}
\caption{NaturalVoices trial counts by task, input-length setting, and
trial condition. Easy positives (EP) pair the same speaker or attribute
within a podcast, whereas hard positives (HP) pair the same speaker or
attribute across podcasts. Easy negatives (EN) pair different speakers
or attributes across podcasts, whereas hard negatives (HN) pair them
within a podcast. All demographic trials are cross-speaker. Here, $P$
is the target probe length in words and $L$ is the maximum number of
enrollment utterances. The four reported settings contain 124{,}356
task-specific trials.}
\label{tab:naturalvoices-trials}
\vspace{-4mm}
\end{table}

\begin{table}[t]
\centering
\tiny
\setlength{\tabcolsep}{5pt}
\begin{tabular}{llrrrrr}
\toprule
\textbf{Probe / enrollment} & \textbf{Task}
& \textbf{EP} & \textbf{HP} & \textbf{EN} & \textbf{HN}
& \textbf{Total} \\
\midrule
\multirow{4}{*}{$P=50,\ L=5$}
 & Speaker & -- & 2,336 & 2,336 & -- & 4,672 \\
 & Sex  & -- & 2,000 & 2,000 & -- & 4,000 \\
 & Age     & -- & 2,000 & 2,000 & -- & 4,000 \\
 & Accent  & -- & 2,000 & 2,000 & -- & 4,000 \\
\midrule
\multirow{4}{*}{$P=50,\ L=20$}
 & Speaker & -- & 2,327 & 2,327 & -- & 4,654 \\
 & Sex  & -- & 2,000 & 2,000 & -- & 4,000 \\
 & Age     & -- & 2,000 & 2,000 & -- & 4,000 \\
 & Accent  & -- & 2,000 & 2,000 & -- & 4,000 \\
\midrule
\multirow{4}{*}{$P=200,\ L=5$}
 & Speaker & -- & 1,970 & 1,970 & -- & 3,940 \\
 & Sex  & -- & 2,000 & 2,000 & -- & 4,000 \\
 & Age     & -- & 2,000 & 2,000 & -- & 4,000 \\
 & Accent  & -- & 2,000 & 2,000 & -- & 4,000 \\
\midrule
\multirow{4}{*}{$P=200,\ L=20$}
 & Speaker & -- & 1,954 & 1,954 & -- & 3,908 \\
 & Sex  & -- & 2,000 & 2,000 & -- & 4,000 \\
 & Age     & -- & 2,000 & 2,000 & -- & 4,000 \\
 & Accent  & -- & 2,000 & 2,000 & -- & 4,000 \\
\bottomrule
\end{tabular}
\caption{SPoRC trial counts by task, input-length setting, and trial
condition. The metadata-filtered SPoRC protocol retains hard-positive
(HP) and easy-negative (EN) trials. Easy-positive (EP) and hard-negative
(HN) conditions were not constructed since the SPoRC podcast has only a single speaker and are marked with dashes.
Positive and negative counts are balanced within every task and setting.
Here, $P$ is the target probe length in words and $L$ is the maximum
number of enrollment utterances. The protocol contains 65{,}174
task-specific trials.}
\label{tab:sporc-trials}
\vspace{-4mm}
\end{table}

\subsection{Model Scoring and Inference Details}
\label{app:inference_details}
\subsubsection{Embedding-based Models}
We evaluate pretrained embedding model baselines without task-specific supervised training: sentence-transformers/all-mpnet-base-v2 (SentenceBERT), AnnaWegmann/Style-Embedding (CISR), Blablablab/multilingual-style-representation (PCM), and rrivera1849/LUAR-CRUD (LUAR). All models are imported from the Sentence Transformers library and run on a single NVIDIA A40 GPU. For SentenceBERT, CISR, and PCM, we average L2-normalized utterance embeddings within each side and compute cosine similarity. TF--IDF uses word unigram and bigram representations fitted without labels to the unique utterances in each task and input configuration. LUAR receives 16 utterances per side: longer sets are truncated, while shorter sets are padded by repeating the final utterance; each utterance is limited to 32 tokens. Probe and enrollment representations are compared using cosine similarity.

\subsubsection{LLM Scoring}
We query gpt-5-mini through the OpenAI API and meta-llama/Llama-3.3-70B-Instruct-Turbo and Qwen/Qwen3.5-9B through the Together API. Each trial is presented as a single zero-shot user message. GPT-5-mini uses low reasoning effort and low response verbosity, while Llama uses the provider's default decoding settings; no temperature is explicitly supplied to either model. For Qwen, we disable reasoning and use temperature (1.0), \texttt{top\_p}=0.95, \texttt{top\_k}=20, \texttt{min\_p}=0, presence penalty (1.5), and repetition penalty (1.0), following the model card's recommended non-thinking configuration for reasoning tasks. These settings were not tuned on the evaluation labels. We extract the first numeric value from each response and constrain it to the requested 0--100 range.

All three LLMs receive the same prompt structure. The identity-verification prompt is:

\begin{quote}
\small
You are a speaker verification judge.\

Estimate how likely it is that the probe and enrollment utterances came from the SAME SPEAKER.\
Focus on stylistic and linguistic cues, not topic similarity.\
Output one number from 0 to 100:\
0 = definitely different speaker\
100 = definitely same speaker\

Do not explain your reasoning.\

Probe utterances:\

[probe utterance 1]\
(\ldots)\

Enrollment utterances:\

[enrollment utterance 1]\
(\ldots)
\end{quote}

For attribute verification, the target and endpoint descriptions are replaced by same biological sex, same age group (<30, 30--60, or 60+), or same accent group (North America versus Other). The sex and age prompts ask the model to focus on stylistic and linguistic cues, whereas the accent prompt specifies vocabulary, phrasing, and linguistic markers. The complete task-specific prompts are included in the released code.

\subsection{Full Results}
\label{app:full-results}

This section reports complete results for all eight systems (Section~5.1) on both corpora, across all trial conditions and the four input-budget settings (probe length $P \in \{50, 200\}$ words, enrollment size $L \in \{5, 20\}$ utterances). The main paper reports the base configuration ($P{=}200$, $L{=}20$) for a representative subset of models; the figures and tables below provide the corresponding results for every system and setting.

\subsubsection{SPoRC}
\label{app:sporc-results}

\begin{figure*}[t]
    \centering
    \begin{subfigure}[b]{0.48\linewidth}
        \includegraphics[width=\linewidth]{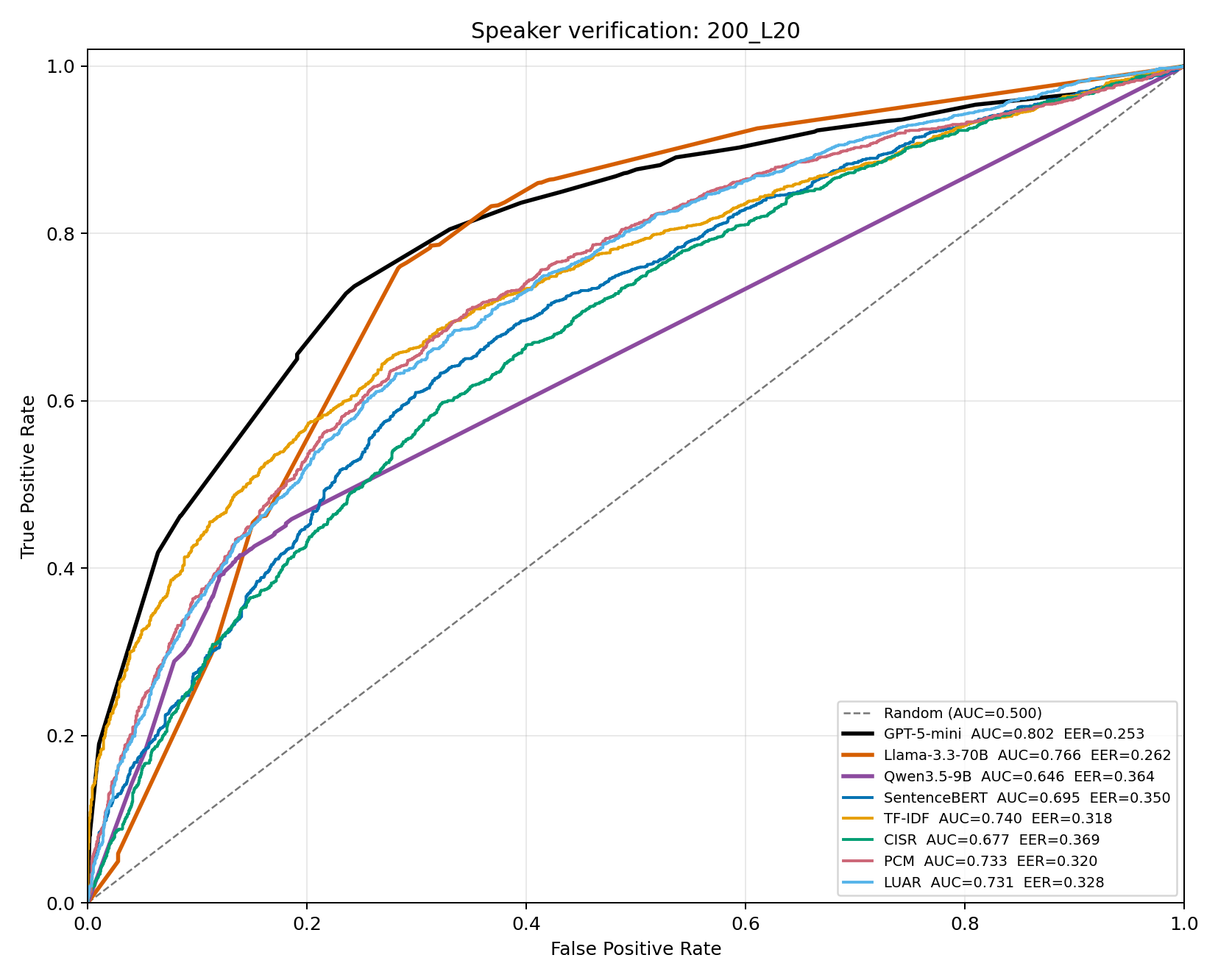}
        \caption{$P=200$, $L=20$}
        \label{fig:sporc_speaker_200_L20}
    \end{subfigure}
    \hfill
    \begin{subfigure}[b]{0.48\linewidth}
        \includegraphics[width=\linewidth]{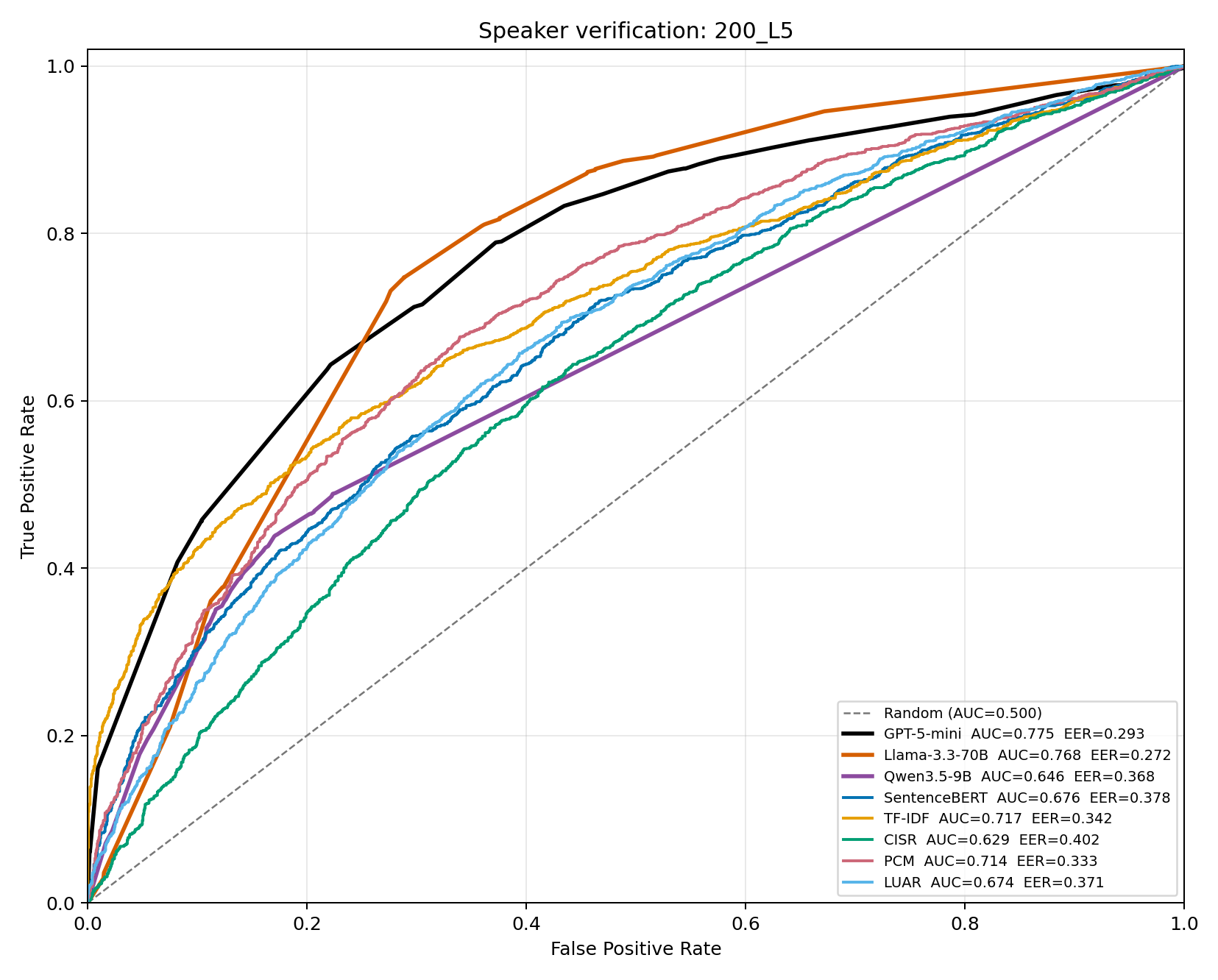}
        \caption{$P=200$, $L=5$}
        \label{fig:sporc_speaker_200_L5}
    \end{subfigure}

    \vspace{0.5em}

    \begin{subfigure}[b]{0.48\linewidth}
        \includegraphics[width=\linewidth]{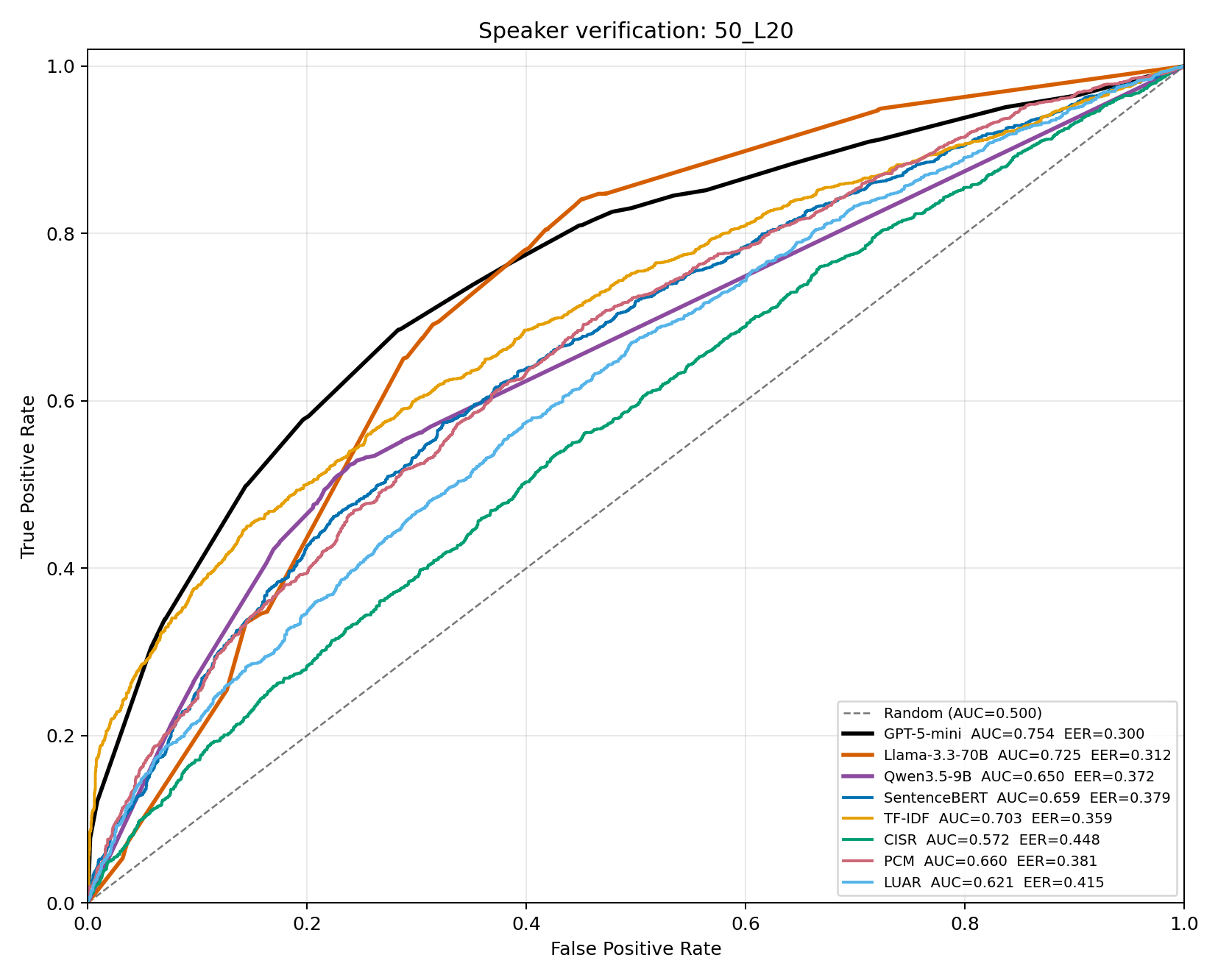}
        \caption{$P=50$, $L=20$}
        \label{fig:sporc_speaker_50_L20}
    \end{subfigure}
    \hfill
    \begin{subfigure}[b]{0.48\linewidth}
        \includegraphics[width=\linewidth]{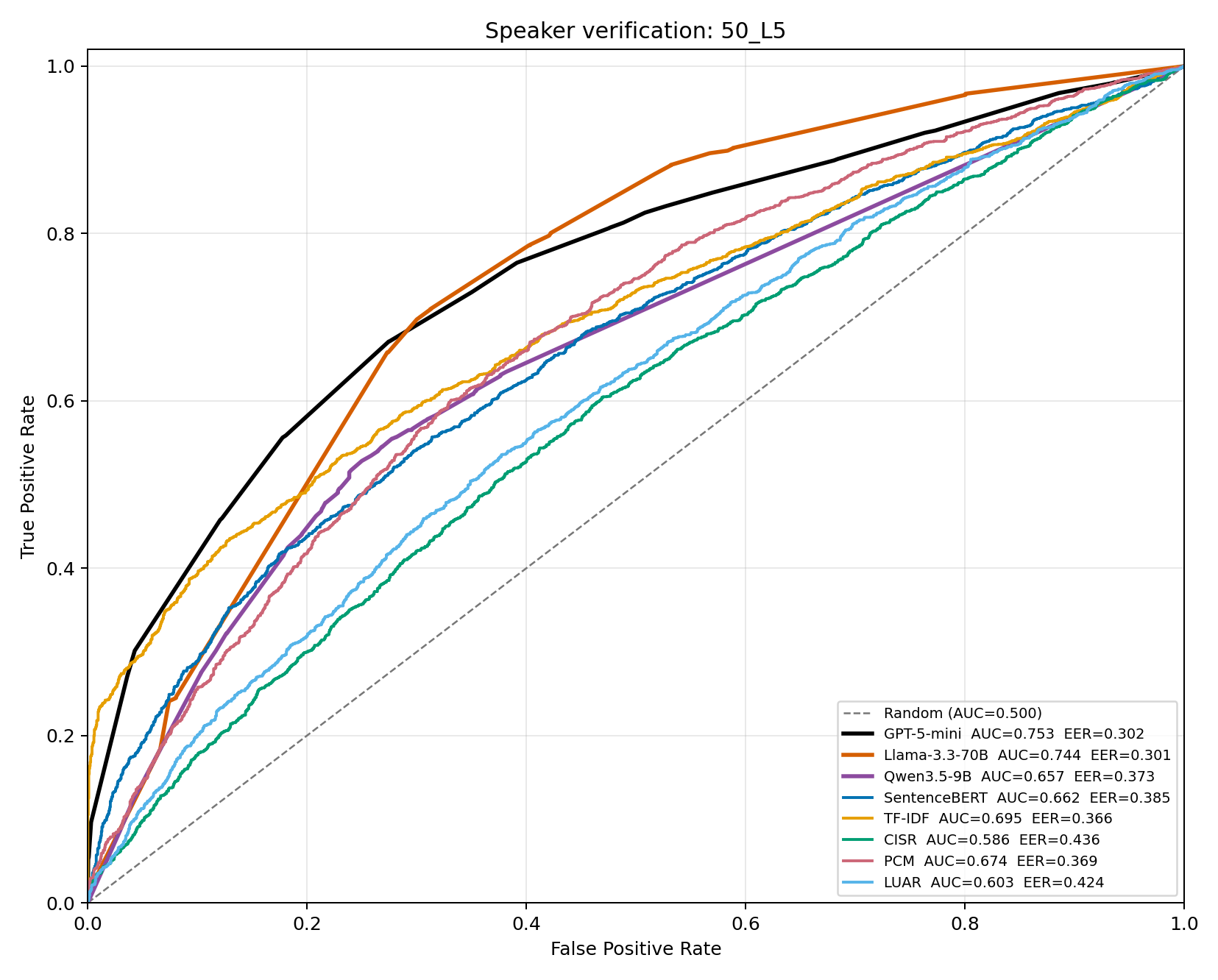}
        \caption{$P=50$, $L=5$}
        \label{fig:sporc_speaker_50_L5}
    \end{subfigure}
    \caption{ROC curves for speaker identity verification on SPoRC
    (different-topic condition) for all eight models, across the four
    probe-length ($P$, in words) and enrollment-size ($L$, in utterances)
    settings.}
    \label{fig:sporc_speaker_roc}
\end{figure*}

\begin{figure*}[t]
    \centering
    \begin{subfigure}[b]{0.48\linewidth}
        \includegraphics[width=\linewidth]{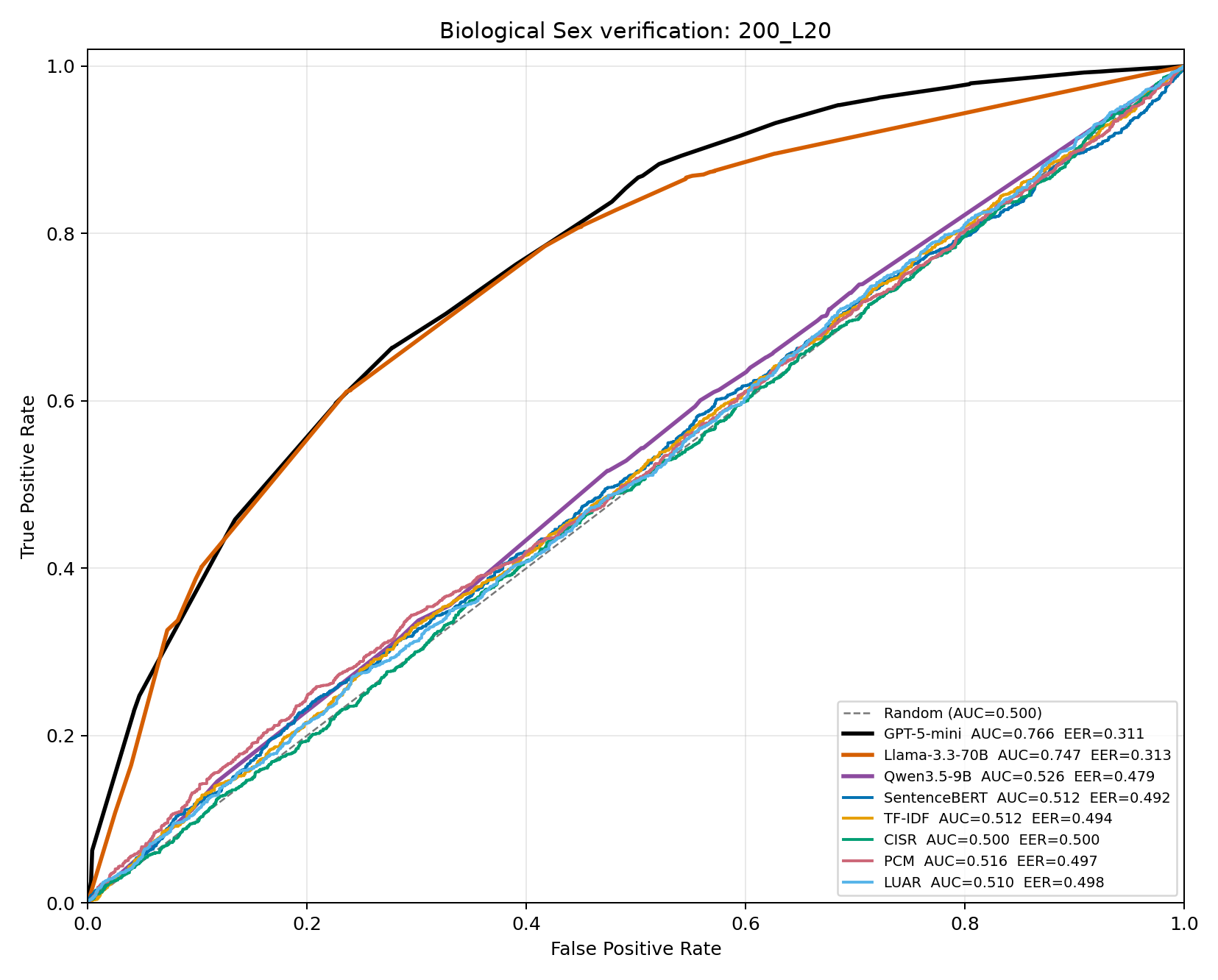}
        \caption{$P=200$, $L=20$}
        \label{fig:sporc_sex_200_L20}
    \end{subfigure}
    \hfill
    \begin{subfigure}[b]{0.48\linewidth}
        \includegraphics[width=\linewidth]{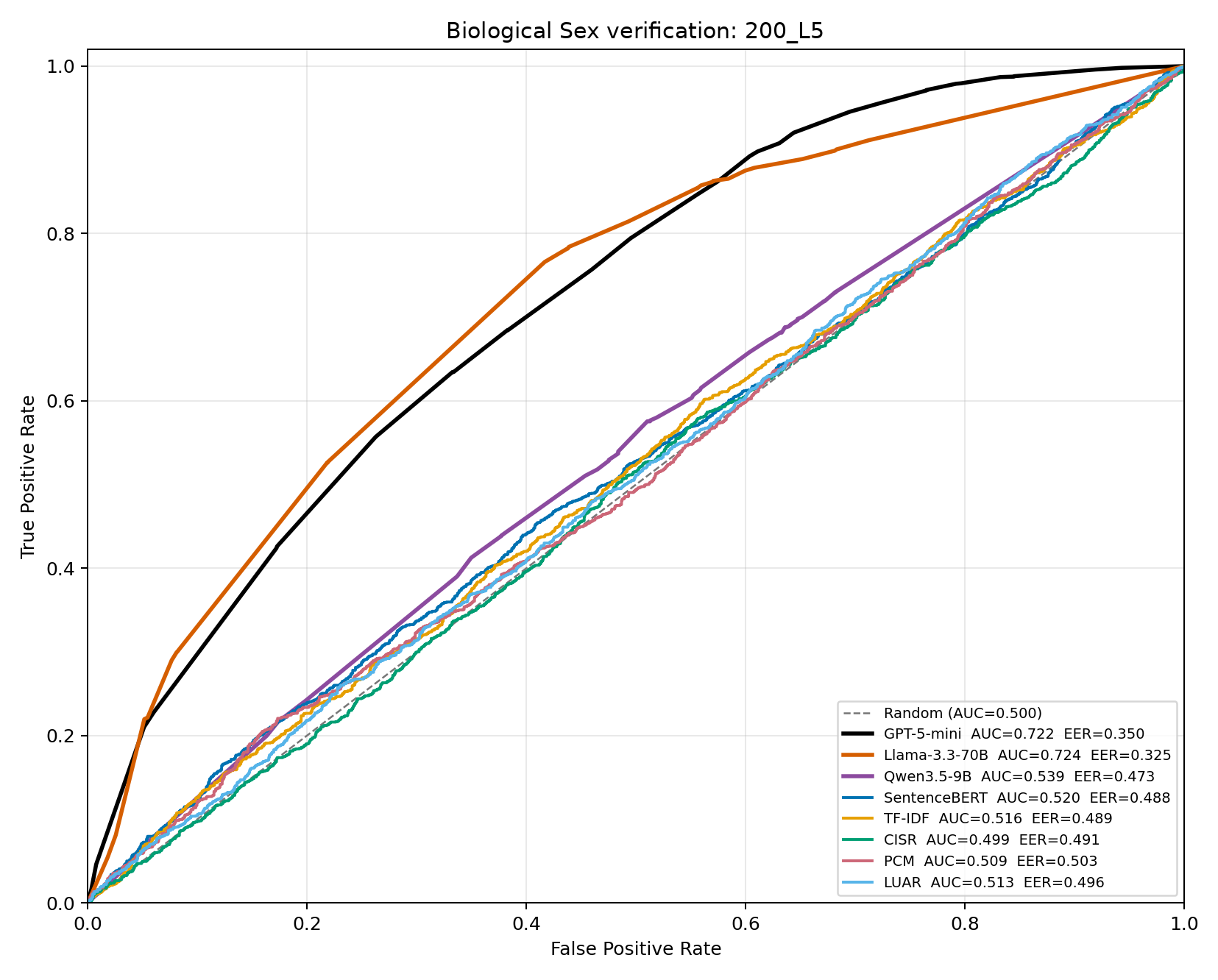}
        \caption{$P=200$, $L=5$}
        \label{fig:sporc_sex_200_L5}
    \end{subfigure}

    \vspace{0.5em}

    \begin{subfigure}[b]{0.48\linewidth}
        \includegraphics[width=\linewidth]{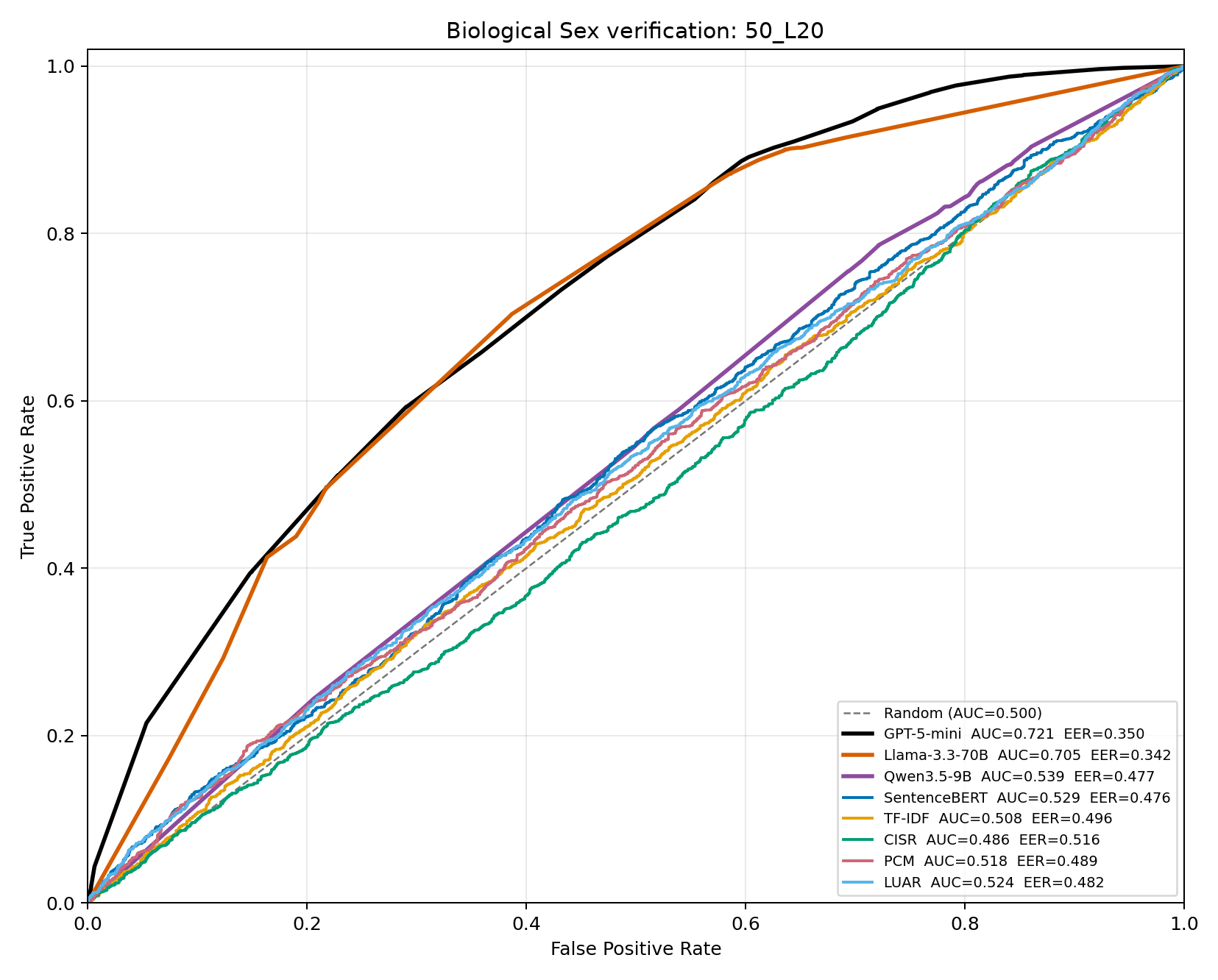}
        \caption{$P=50$, $L=20$}
        \label{fig:sporc_sex_50_L20}
    \end{subfigure}
    \hfill
    \begin{subfigure}[b]{0.48\linewidth}
        \includegraphics[width=\linewidth]{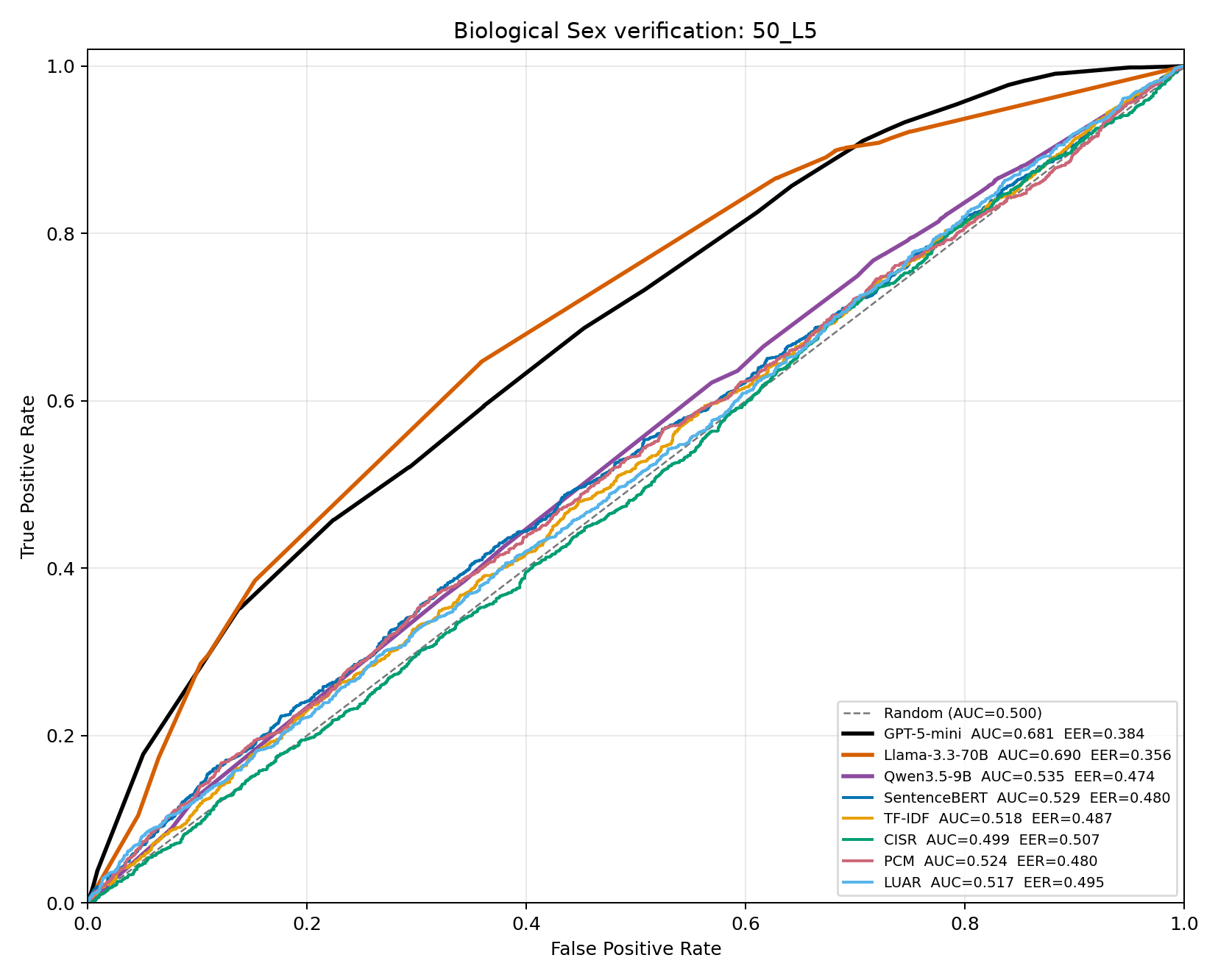}
        \caption{$P=50$, $L=5$}
        \label{fig:sporc_sex_50_L5}
    \end{subfigure}
\caption{ROC curves for sex-label verification on SPoRC (different-topic
    condition) for all eight models, across the four probe-length ($P$)
    and enrollment-size ($L$) settings.}
    \label{fig:sporc_sex_roc}
\end{figure*}

\begin{figure*}[t]
    \centering
    \begin{subfigure}[b]{0.48\linewidth}
        \includegraphics[width=\linewidth]{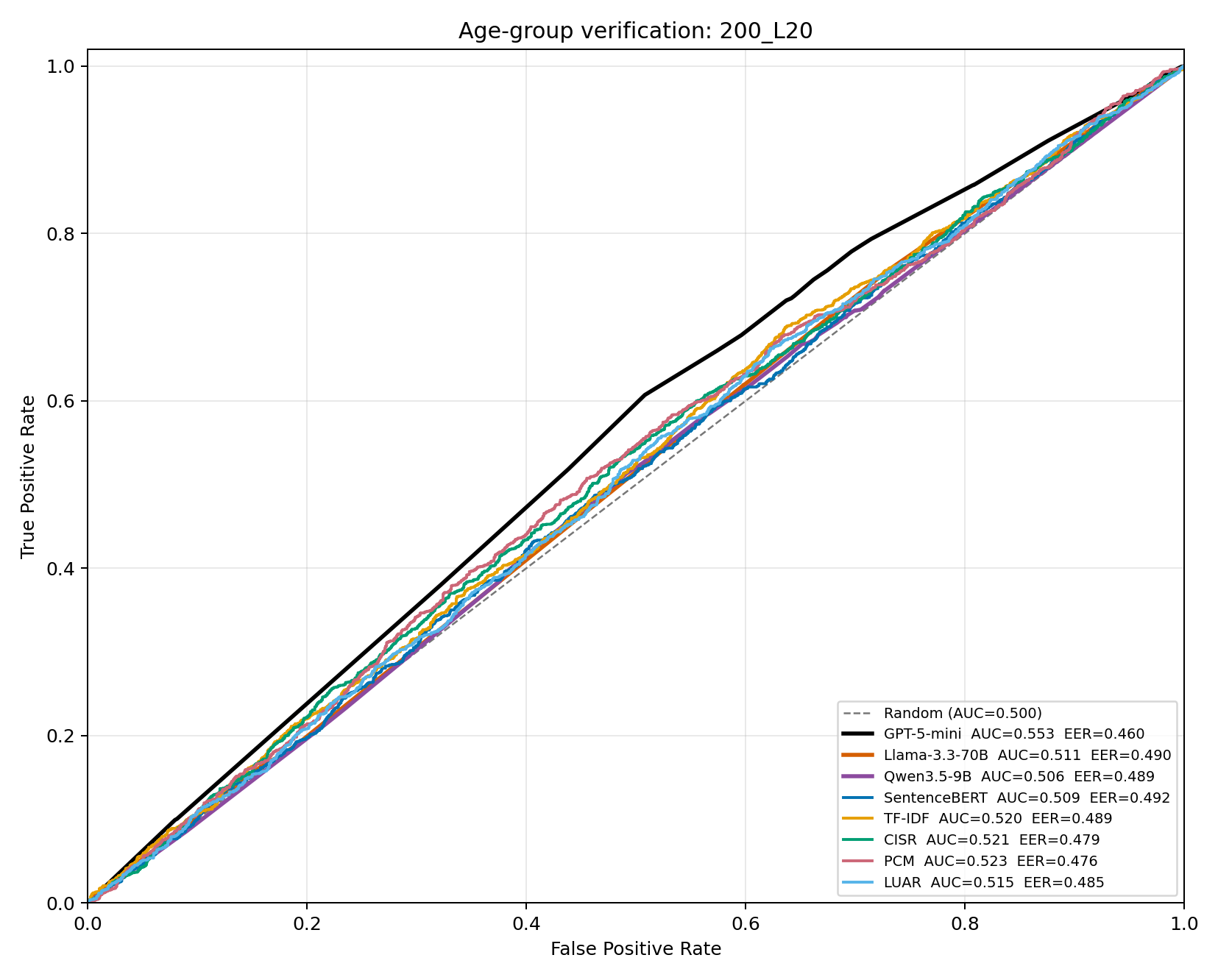}
        \caption{$P=200$, $L=20$}
        \label{fig:sporc_age_200_L20}
    \end{subfigure}
    \hfill
    \begin{subfigure}[b]{0.48\linewidth}
        \includegraphics[width=\linewidth]{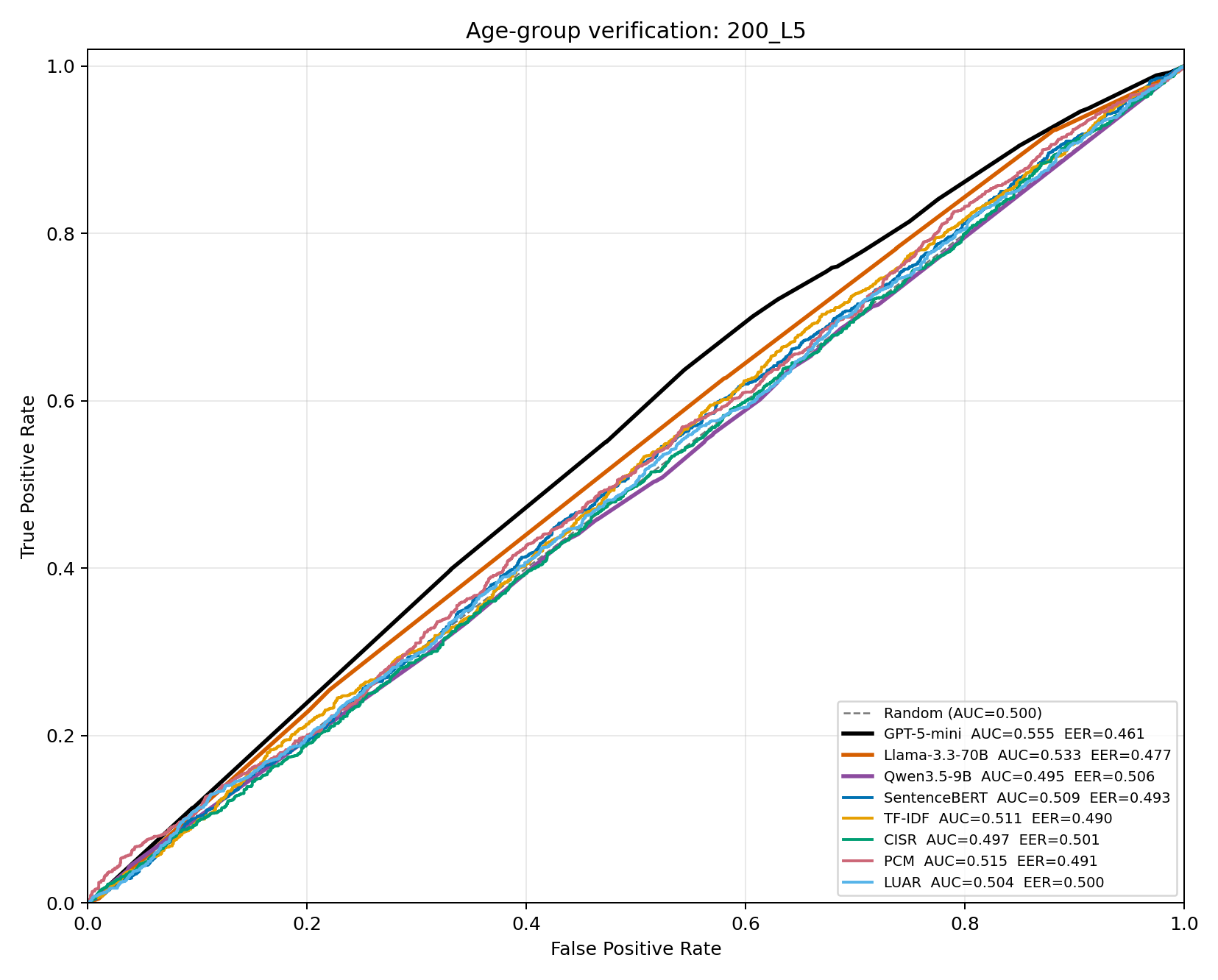}
        \caption{$P=200$, $L=5$}
        \label{fig:sporc_age_200_L5}
    \end{subfigure}

    \vspace{0.5em}

    \begin{subfigure}[b]{0.48\linewidth}
        \includegraphics[width=\linewidth]{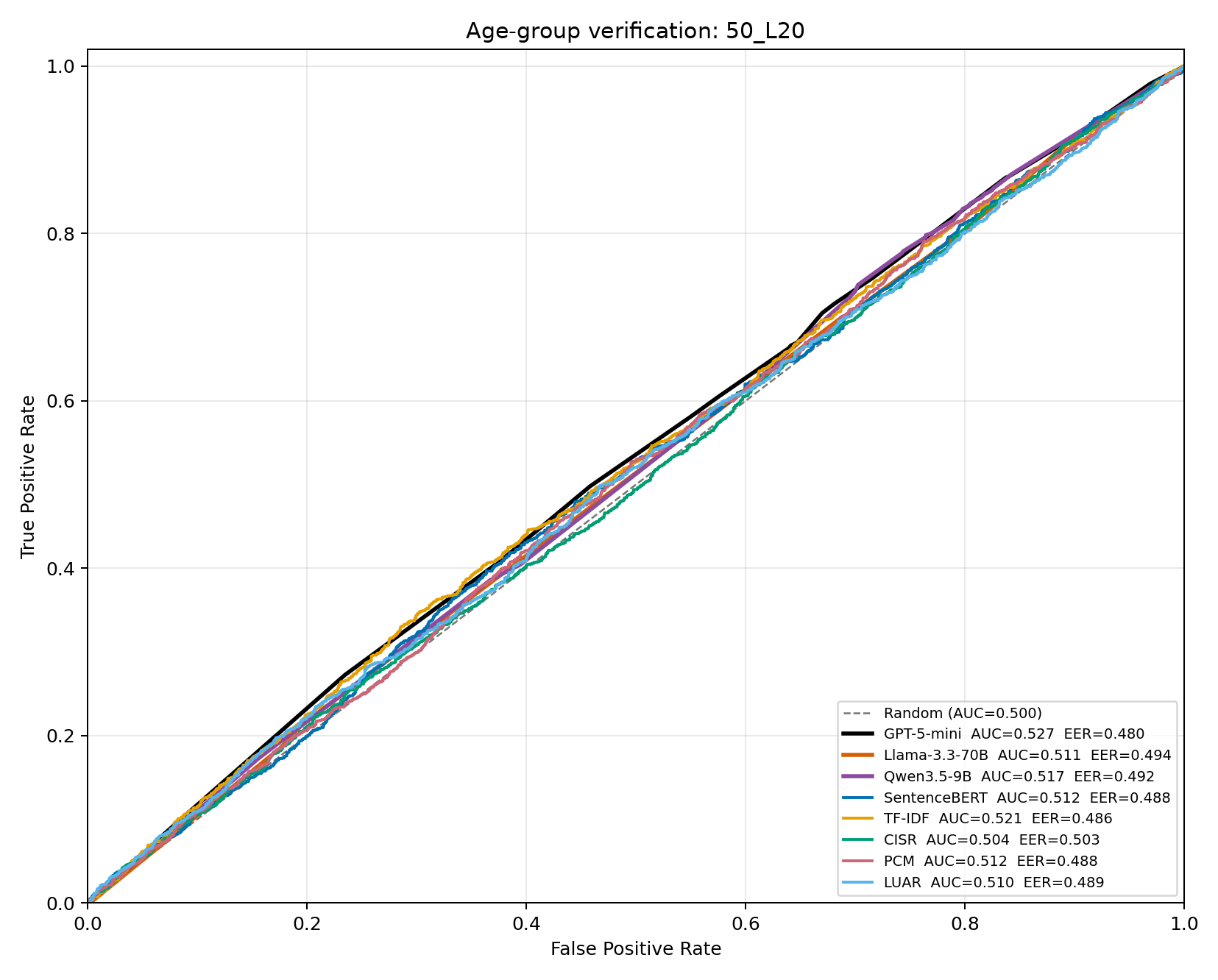}
        \caption{$P=50$, $L=20$}
        \label{fig:sporc_age_50_L20}
    \end{subfigure}
    \hfill
    \begin{subfigure}[b]{0.48\linewidth}
        \includegraphics[width=\linewidth]{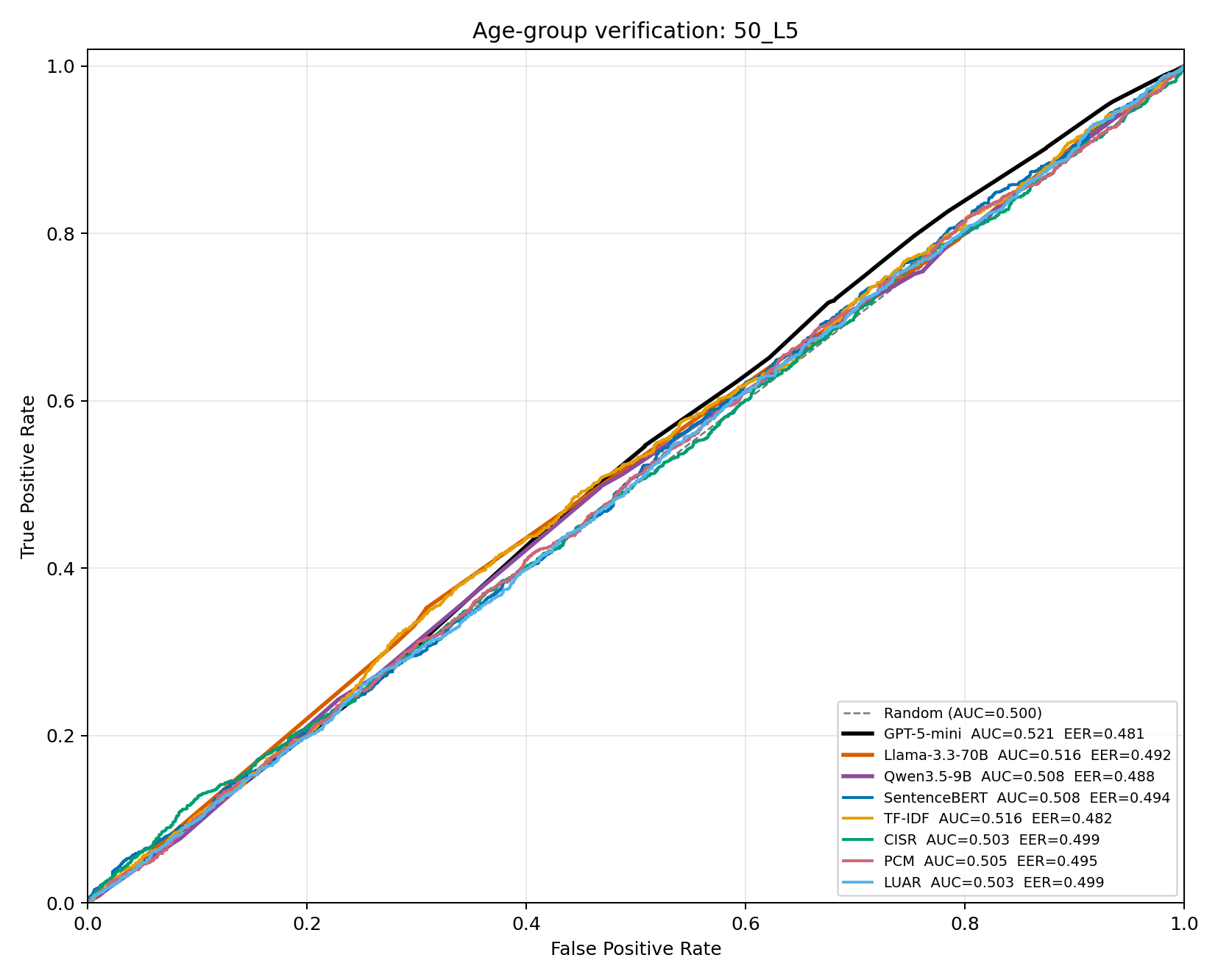}
        \caption{$P=50$, $L=5$}
        \label{fig:sporc_age_50_L5}
    \end{subfigure}
    \caption{ROC curves for age-group verification on SPoRC
    (different-topic condition) for all eight models, across the four
    probe-length ($P$) and enrollment-size ($L$) settings.}
    \label{fig:sporc_age_roc}
\end{figure*}

\begin{figure*}[t]
    \centering
    \begin{subfigure}[b]{0.48\linewidth}
        \includegraphics[width=\linewidth]{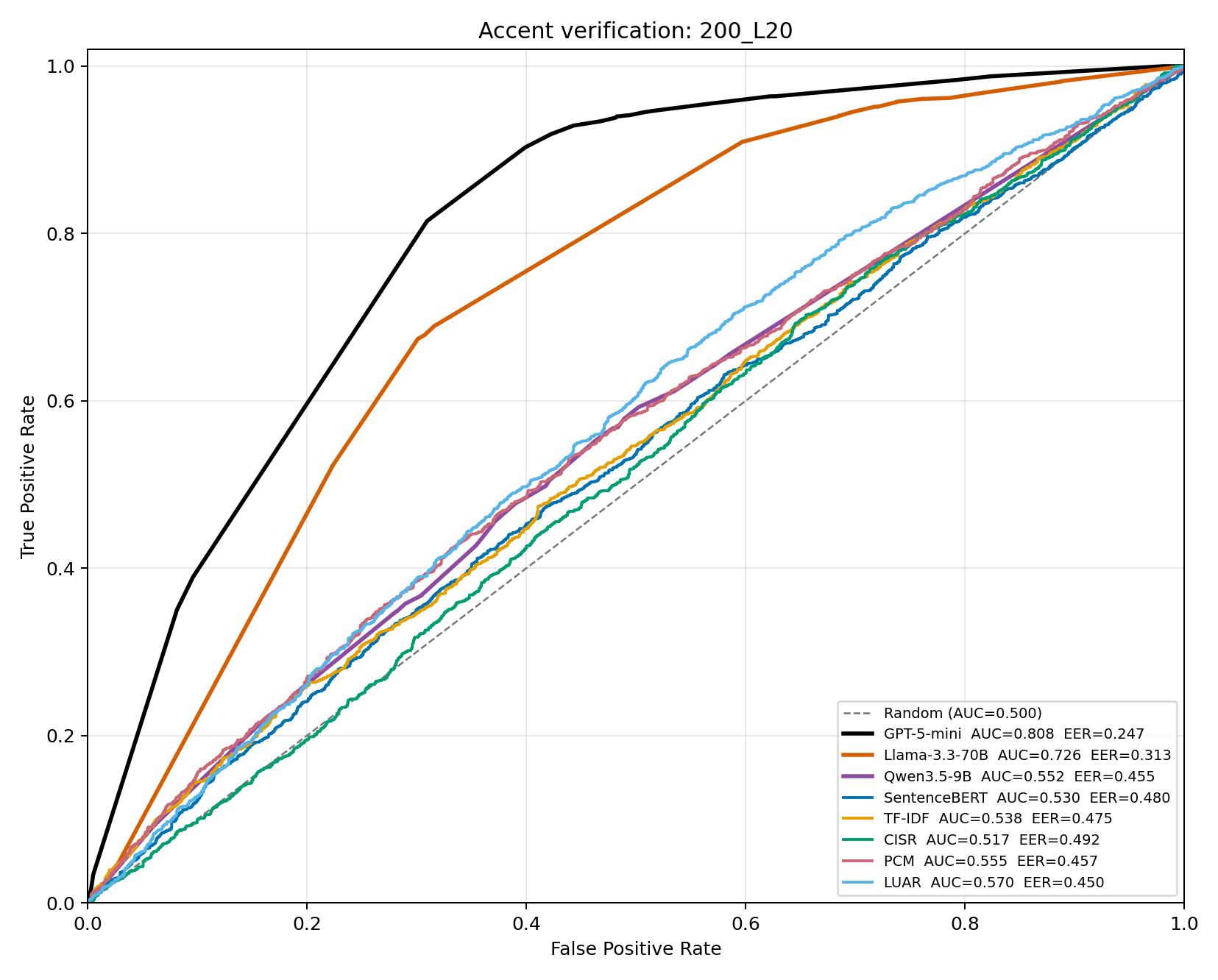}
        \caption{$P=200$, $L=20$}
        \label{fig:sporc_accent_200_L20}
    \end{subfigure}
    \hfill
    \begin{subfigure}[b]{0.48\linewidth}
        \includegraphics[width=\linewidth]{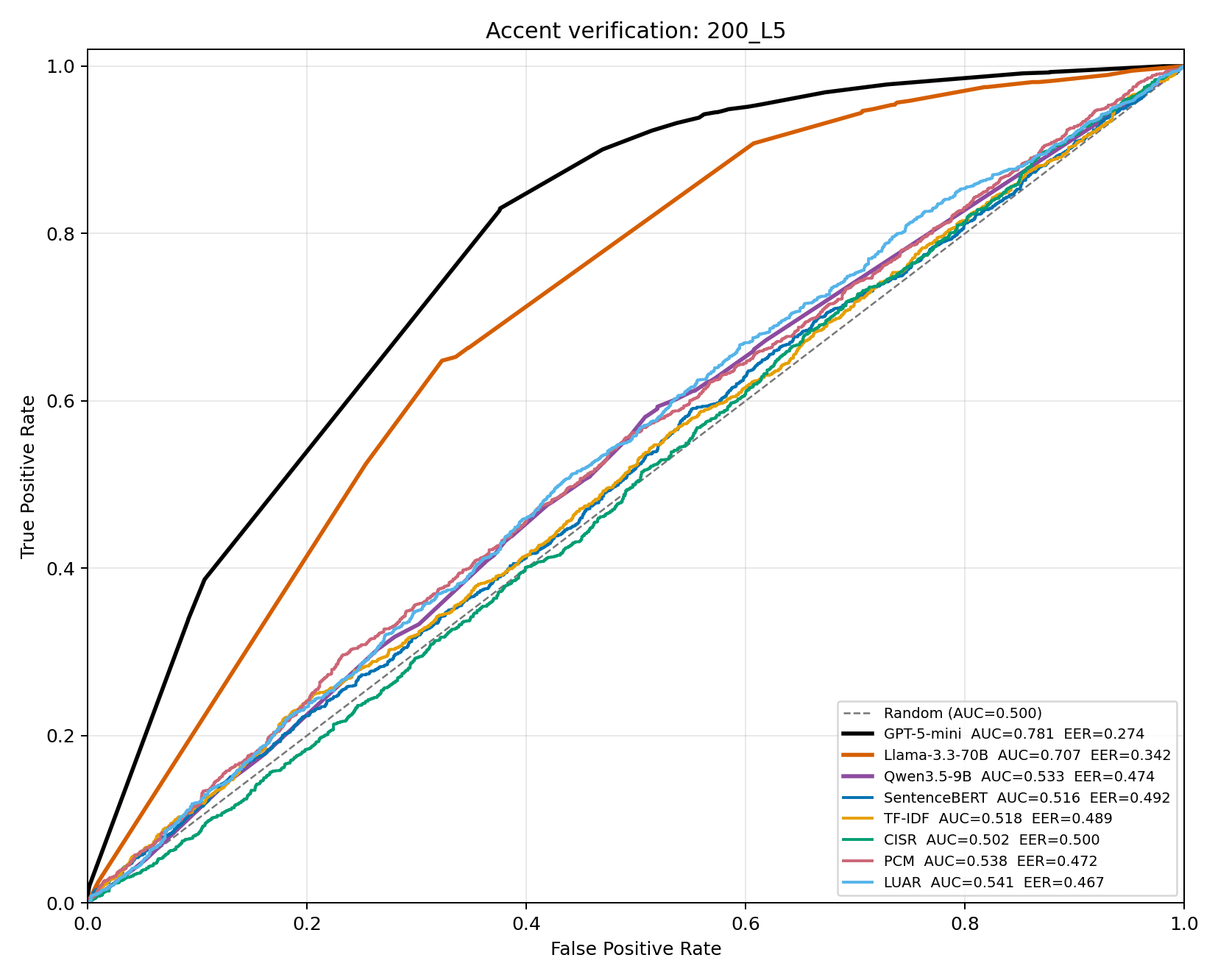}
        \caption{$P=200$, $L=5$}
        \label{fig:sporc_accent_200_L5}
    \end{subfigure}

    \vspace{0.5em}

    \begin{subfigure}[b]{0.48\linewidth}
        \includegraphics[width=\linewidth]{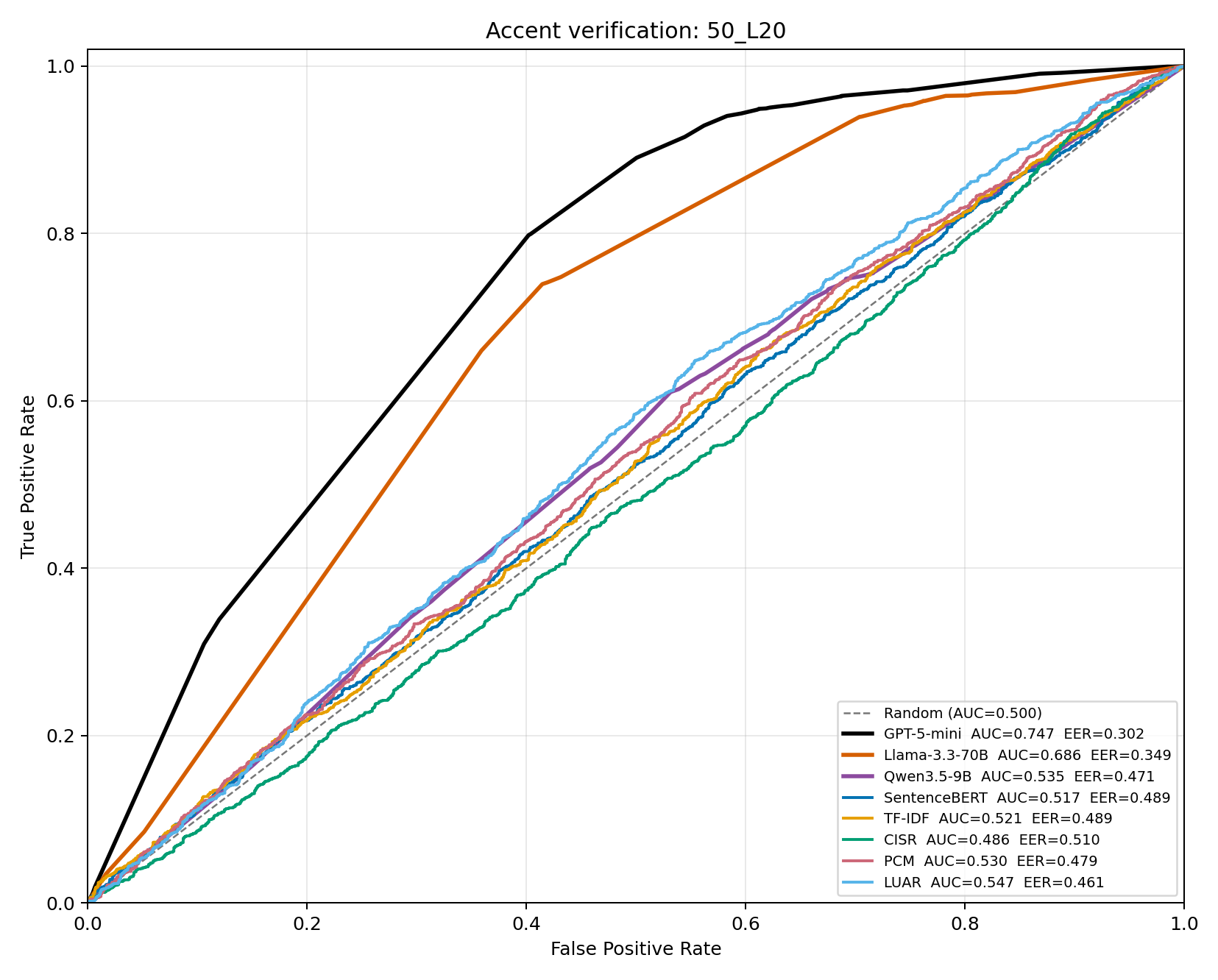}
        \caption{$P=50$, $L=20$}
        \label{fig:sporc_accent_50_L20}
    \end{subfigure}
    \hfill
    \begin{subfigure}[b]{0.48\linewidth}
        \includegraphics[width=\linewidth]{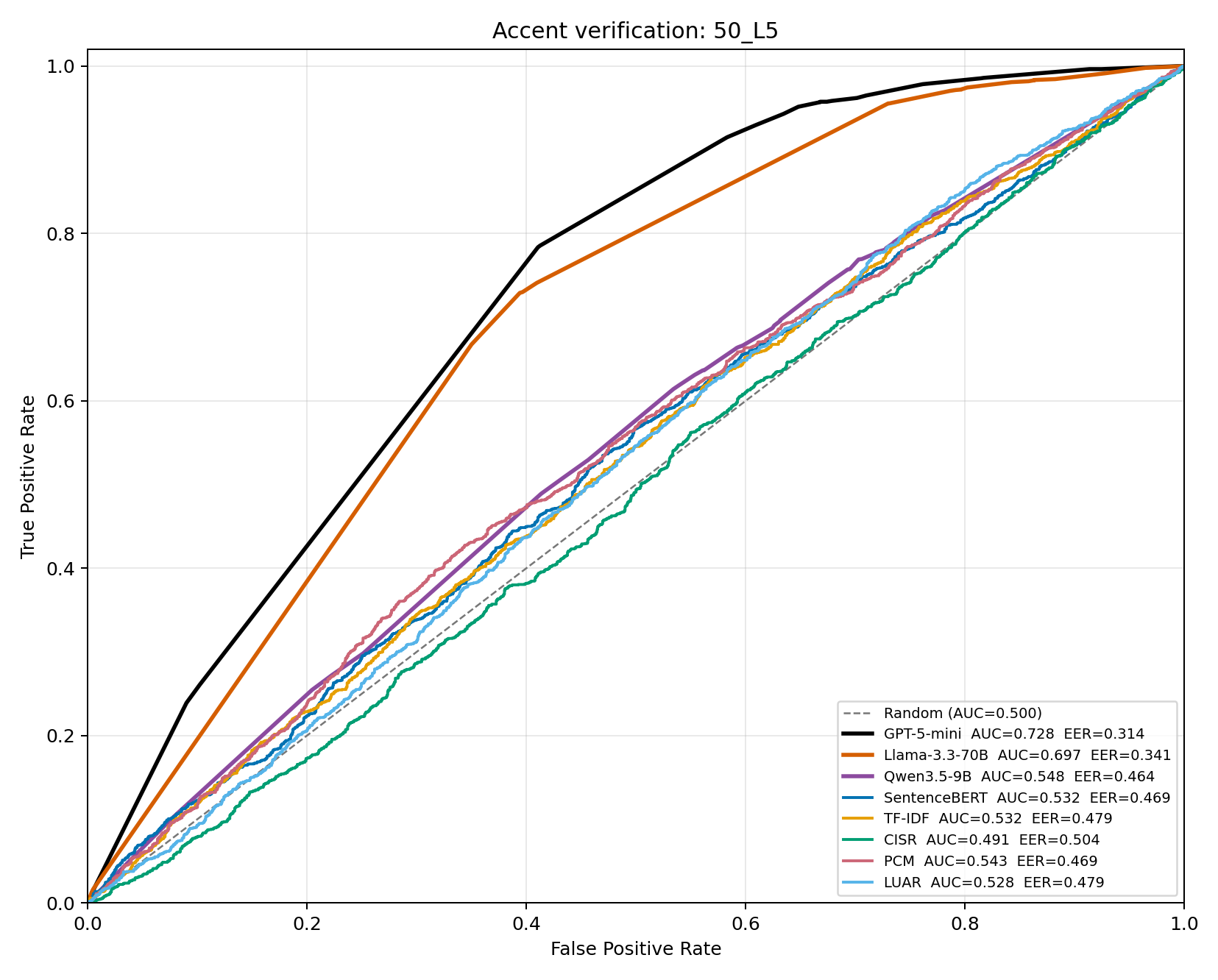}
        \caption{$P=50$, $L=5$}
        \label{fig:sporc_accent_50_L5}
    \end{subfigure}
    \caption{ROC curves for accent verification on SPoRC (different-topic
    condition) for all eight models, across the four probe-length ($P$)
    and enrollment-size ($L$) settings.}
    \label{fig:sporc_accent_roc}
\end{figure*}

Figures~\ref{fig:sporc_speaker_roc}-\ref{fig:sporc_accent_roc} show ROC curves for the four SPoRC verification tasks under each of the four input-budget settings. SPoRC supports only cross-episode pairs (Section~\ref{trial_gen}), with metadata-based topic proxies and a series-membership imbalance in the original speaker protocol. The two strongest LLMs outperform the remaining systems on identity verification; their advantage on sex- and accent-label agreement persists while age-label agreement remains near chance. Performance generally degrades with shorter probes, whereas reduced enrollment has a smaller effect.

\subsubsection{NaturalVoices}
\label{app:nv-results}

Tables~\ref{tab:speaker_full}--\ref{tab:accent_full} report EER and AUC for all eight systems on NaturalVoices, covering the combined, easy, hard, same-topic, and different-topic conditions, with $\Delta$ columns giving the change from the base configuration under each reduced input budget.

The easy/hard split, omitted from the main text for space, shows the effect of episode co-membership. On easy trials, probe and enrollment share an episode only on positive pairs, and SentenceBERT reaches 0.168 EER on identity. On hard trials, episode co-membership points toward the wrong label, and the embedding baselines have EERs of 0.546--0.621 on identity, consistent with reliance on episode context. The LLMs also degrade (GPT-5-mini: 0.542 EER; Llama-3.3-70B: 0.556). The same pattern appears for demographic-label agreement, where hard-condition EERs for the embedding models reach 0.62--0.80.

The ablation columns extend the analysis of Section~\ref{sec:ablations} to all systems and conditions. The probe-length effect is consistent across model families. In the cross-episode condition, reducing enrollment from $L=20$ to $L=5$ generally costs less than shortening the probe. Qwen3.5-9B tracks the embedding baselines throughout (0.418 EER on cross-episode identity, chance-level attribute performance), indicating that the LLM advantage observed in the main text is not shared by every LLM. Finally, the age-label results for Llama-3.3-70B (0.546 combined EER) improve under reduced information, consistent with response bias rather than a reliable age signal.

\subsection{Cross-Series SPoRC Speaker Verification}
\label{app:cross-series}
The original SPoRC speaker protocol includes 1,594 cross-series positive trials across the four input settings, while every negative trial crosses series (Appendix~\ref{app:trial-construction}). Restricting both classes to cross-series pairs removes series co-membership as a direct class cue. Table~\ref{tab:sporc-cross-series} reports EER for the two strongest LLMs. At the base configuration, EER rises from 0.253 to 0.402 for GPT-5-mini and from 0.262 to 0.391 for Llama-3.3-70B, but remains below the 0.5 chance value. Under this control, EER varies by at most 0.011 (GPT-5-mini) and 0.015 (Llama-3.3-70B) across the four input settings, so the probe-length effect of Section~\ref{sec:ablations} is not visible in this stricter protocol. This control does not eliminate all differences in semantic topic or speaker-specific context.

\begin{table}[h]
\centering
\small
\begin{tabular}{lcc}
\toprule
\textbf{Probe/enrollment} & \textbf{GPT-5-mini} & \textbf{Llama-3.3-70B} \\
\midrule
$P=50,\ L=5$   & 0.410 & 0.390 \\
$P=50,\ L=20$  & 0.406 & 0.376 \\
$P=200,\ L=5$  & 0.399 & 0.390 \\
$P=200,\ L=20$ & 0.402 & 0.391 \\
\bottomrule
\end{tabular}
\caption{EER for SPoRC speaker verification when positive and negative pairs both cross podcast series (lower is better).}
\label{tab:sporc-cross-series}
\vspace{-4mm}
\end{table}

\subsection{Human Audit of Inferred Attribute Labels}
\label{app:human-validation}
Three independent annotators judged 50 recordings per attribute, sampled to balance the categories of that attribute. Table~\ref{tab:human-validation} compares the model-inferred labels with the annotator majority and reports inter-annotator agreement. The audit measures perceptual agreement on this selected sample rather than verified demographic ground truth or corpus-wide accuracy. Agreement on age group is notably lower than for the other two attributes.

\begin{table}[h]
\centering
\small
\begin{tabular}{lccc}
\toprule
\textbf{Metric} & \textbf{Age group} & \textbf{Sex} & \textbf{Accent} \\
\midrule
Human agr. & 60--76\% & 94--98\% & 80--82\% \\
Model--majority agr. & 64\% & 98\% & 88\% \\
Human $\kappa$ & 0.489 & 0.920 & 0.613 \\
Model--majority $\kappa$ & 0.531 & 0.960 & 0.760 \\
\bottomrule
\end{tabular}
\caption{Perceptual audit of model-inferred attribute labels on 50 category-balanced recordings per attribute. Human agreement is shown as the reported range.}
\label{tab:human-validation}
\end{table}

\begin{table*}[h]
\centering
\small
\setlength{\tabcolsep}{3.5pt}
\begin{tabular}{ll rr rr rr rr}
\toprule
& & \multicolumn{2}{c}{\textbf{Main - 200w / }$L$\textbf{=20}}
  & \multicolumn{2}{c}{$\Delta$ \textbf{200w / }$L$\textbf{=5}}
  & \multicolumn{2}{c}{$\Delta$ \textbf{50w / }$L$\textbf{=20}}
  & \multicolumn{2}{c}{$\Delta$ \textbf{50w / }$L$\textbf{=5}} \\
\cmidrule(lr){3-4}\cmidrule(lr){5-6}\cmidrule(lr){7-8}\cmidrule(lr){9-10}
\textbf{Model} & \textbf{Condition}
  & EER\,$(\downarrow)$ & AUC\,$(\uparrow)$
  & EER\,$(\downarrow)$ & AUC\,$(\uparrow)$
  & EER\,$(\downarrow)$ & AUC\,$(\uparrow)$
  & EER\,$(\downarrow)$ & AUC\,$(\uparrow)$\\
\midrule
\multirow{5}{*}{SentenceBERT}
  & Combined   & 0.350 & 0.683 & \deer{ 0.048} & \dauc{-0.049} & \deer{ 0.043} & \dauc{-0.046} & \deer{ 0.062} & \dauc{-0.062} \\
  & Easy       & 0.168 & 0.913 & \deer{ 0.071} & \dauc{-0.072} & \deer{ 0.070} & \dauc{-0.071} & \deer{ 0.110} & \dauc{-0.118} \\
  & Hard       & 0.618 & 0.334 & \deer{-0.005} & \dauc{ 0.017} & \deer{-0.009} & \dauc{ 0.020} & \deer{-0.032} & \dauc{ 0.050} \\
  & Same Topic & 0.460 & 0.567 & \deer{ 0.040} & \dauc{-0.064} & \deer{ 0.004} & \dauc{-0.025} & \deer{ 0.020} & \dauc{-0.034} \\
  & Diff Topic & 0.351 & 0.708 & \deer{0.024} & \dauc{-0.025} & \deer{0.038} & \dauc{-0.047} & \deer{0.046} & \dauc{-0.059}
  \\
\midrule
\multirow{5}{*}{TF-IDF}
  & Combined   & 0.362 & 0.678 & \deer{ 0.061} & \dauc{-0.069} & \deer{ 0.064} & \dauc{-0.074} & \deer{ 0.078} & \dauc{-0.087} \\
  & Easy       & 0.229 & 0.852 & \deer{ 0.103} & \dauc{-0.116} & \deer{ 0.101} & \dauc{-0.115} & \deer{ 0.142} & \dauc{-0.163} \\
  & Hard       & 0.546 & 0.439 & \deer{ 0.006} & \dauc{ 0.003} & \deer{ 0.008} & \dauc{-0.004} & \deer{-0.014} & \dauc{ 0.027} \\
  & Same Topic & 0.409 & 0.637 & \deer{ 0.045} & \dauc{-0.073} & \deer{ 0.055} & \dauc{-0.075} & \deer{ 0.044} & \dauc{-0.077} \\
  & Diff Topic & 0.404 & 0.644 & \deer{ 0.027} & \dauc{-0.039} & \deer{ 0.034} & \dauc{-0.051} & \deer{ 0.047} & \dauc{-0.059} \\
\midrule
\multirow{5}{*}{CISR}
  & Combined   & 0.441 & 0.580 & \deer{ 0.022} & \dauc{-0.036} & \deer{ 0.026} & \dauc{-0.032} & \deer{ 0.029} & \dauc{-0.041} \\
  & Easy       & 0.313 & 0.755 & \deer{ 0.041} & \dauc{-0.062} & \deer{ 0.050} & \dauc{-0.061} & \deer{ 0.066} & \dauc{-0.093} \\
  & Hard       & 0.621 & 0.329 & \deer{-0.013} & \dauc{ 0.012} & \deer{-0.015} & \dauc{ 0.021} & \deer{-0.038} & \dauc{ 0.044} \\
  & Same Topic & 0.492 & 0.505 & \deer{ 0.027} & \dauc{-0.042} & \deer{ 0.021} & \dauc{-0.027} & \deer{ 0.026} & \dauc{-0.038} \\
  & Diff Topic & 0.441 & 0.579 & \deer{ 0.007} & \dauc{-0.009} & \deer{ 0.015} & \dauc{-0.015} & \deer{ 0.016} & \dauc{-0.014} \\
\midrule
\multirow{5}{*}{PCM}
  & Combined   & 0.386 & 0.650 & \deer{ 0.018} & \dauc{-0.022} & \deer{ 0.040} & \dauc{-0.048} & \deer{ 0.077} & \dauc{-0.098} \\
  & Easy       & 0.224 & 0.854 & \deer{ 0.034} & \dauc{-0.037} & \deer{ 0.075} & \dauc{-0.082} & \deer{ 0.075} & \dauc{-0.080} \\
  & Hard       & 0.618 & 0.348 & \deer{-0.027} & \dauc{ 0.023} & \deer{-0.027} & \dauc{ 0.027} & \deer{-0.055} & \dauc{ 0.059} \\
  & Same Topic & 0.462 & 0.555 & \deer{ 0.018} & \dauc{-0.029} & \deer{ 0.028} & \dauc{-0.039} & \deer{-0.002} & \dauc{-0.001} \\
  & Diff Topic & 0.384 & 0.663 & \deer{-0.001} & \dauc{-0.002} & \deer{ 0.030} & \dauc{-0.036} & \deer{ 0.016} & \dauc{-0.027} \\
\midrule
\multirow{5}{*}{LUAR}
  & Combined   & 0.378 & 0.659 & \deer{ 0.035} & \dauc{-0.042} & \deer{ 0.044} & \dauc{-0.055} & \deer{ 0.055} & \dauc{-0.064} \\
  & Easy       & 0.221 & 0.862 & \deer{ 0.061} & \dauc{-0.070} & \deer{ 0.088} & \dauc{-0.098} & \deer{ 0.118} & \dauc{-0.138} \\
  & Hard       & 0.610 & 0.368 & \deer{-0.022} & \dauc{ 0.010} & \deer{-0.025} & \dauc{ 0.027} & \deer{-0.049} & \dauc{ 0.055} \\
  & Same Topic & 0.452 & 0.579 & \deer{ 0.016} & \dauc{-0.035} & \deer{ 0.032} & \dauc{-0.043} & \deer{ 0.014} & \dauc{-0.025} \\
  & Diff Topic & 0.395 & 0.654 & \deer{ 0.016} & \dauc{-0.025} & \deer{ 0.024} & \dauc{-0.042} & \deer{ 0.043} & \dauc{-0.061} \\
\midrule
\multirow{5}{*}{GPT-5-mini}
  & Combined   & \textbf{0.277} & \textbf{0.779} & \deer{ 0.049} & \dauc{-0.052} & \deer{ 0.066} & \dauc{-0.072} & \deer{ 0.082} & \dauc{-0.086} \\
  & Easy       & 0.090 & \textbf{0.964} & \deer{ 0.058} & \dauc{-0.036} & \deer{ 0.088} & \dauc{-0.063} & \deer{ 0.121} & \dauc{-0.095} \\
  & Hard       & \textbf{0.542} & \textbf{0.467} & \deer{ 0.023} & \dauc{-0.041} & \deer{ 0.026} & \dauc{-0.049} & \deer{ 0.014} & \dauc{-0.030} \\
  & Same Topic & 0.345 & \textbf{0.721} & \deer{ 0.044} & \dauc{-0.056} & \deer{ 0.046} & \dauc{-0.065} & \deer{ 0.058} & \dauc{-0.083} \\
  & Diff Topic & 0.285 & 0.790 & \deer{0.041} & \dauc{-0.049} & \deer{0.065} & \dauc{-0.077} & \deer{0.076} & \dauc{-0.089} \\
\midrule
\multirow{5}{*}{Llama-3.3-70B}
  & Combined   & 0.284 & 0.774 & \deer{ 0.042} & \dauc{-0.056} & \deer{ 0.053} & \dauc{-0.061} & \deer{ 0.075} & \dauc{-0.087} \\
  & Easy       & \textbf{0.087} & 0.954 & \deer{ 0.055} & \dauc{-0.035} & \deer{ 0.070} & \dauc{-0.044} & \deer{ 0.116} & \dauc{-0.086} \\
  & Hard       & 0.556 & 0.461 & \deer{ 0.018} & \dauc{-0.044} & \deer{ 0.020} & \dauc{-0.042} & \deer{ 0.014} & \dauc{-0.031} \\
  & Same Topic & \textbf{0.322} & \textbf{0.721} & \deer{ 0.084} & \dauc{-0.078} & \deer{ 0.084} & \dauc{-0.079} & \deer{ 0.110} & \dauc{-0.116} \\
  & Diff Topic & 0.281 & 0.789 & \deer{0.044} & \dauc{-0.045} & \deer{0.047} & \dauc{-0.053} & \deer{0.064} & \dauc{-0.075} \\
\midrule
\multirow{5}{*}{Qwen3.5-9B}
  & Combined   & 0.387 & 0.649 & \deer{ 0.036} & \dauc{-0.044} & \deer{ 0.025} & \dauc{-0.042} & \deer{ 0.037} & \dauc{-0.052} \\
  & Easy       & 0.247 & 0.827 & \deer{ 0.066} & \dauc{-0.065} & \deer{ 0.054} & \dauc{-0.079} & \deer{ 0.071} & \dauc{-0.102} \\
  & Hard       & 0.605 & 0.384 & \deer{-0.012} & \dauc{ 0.001} & \deer{-0.027} & \dauc{ 0.028} & \deer{-0.033} & \dauc{ 0.038} \\
  & Same Topic & 0.418 & 0.598 & \deer{ 0.049} & \dauc{-0.053} & \deer{ 0.035} & \dauc{-0.041} & \deer{ 0.055} & \dauc{-0.059} \\
  & Diff Topic & 0.418 & 0.635 & \deer{ 0.013} & \dauc{-0.021} & \deer{ 0.007} & \dauc{-0.025} & \deer{ 0.004} & \dauc{-0.026} \\
\bottomrule
\end{tabular}
\caption{Identity verification results on NaturalVoices for all models across trial conditions and ablation settings.
Main results use probe\,$\approx$\,200 words and $L=20$;
$\Delta$ columns show the change from the main setting.
Best EER and AUC per condition (main setting) in \textbf{bold}.}
\label{tab:speaker_full}
\end{table*}

\begin{table*}[h]
\centering
\small
\setlength{\tabcolsep}{3.5pt}
\begin{tabular}{ll rr rr rr rr}
\toprule
& & \multicolumn{2}{c}{\textbf{Main - 200w / }$L$\textbf{=20}}
  & \multicolumn{2}{c}{$\Delta$ \textbf{200w / }$L$\textbf{=5}}
  & \multicolumn{2}{c}{$\Delta$ \textbf{50w / }$L$\textbf{=20}}
  & \multicolumn{2}{c}{$\Delta$ \textbf{50w / }$L$\textbf{=5}} \\
\cmidrule(lr){3-4}\cmidrule(lr){5-6}\cmidrule(lr){7-8}\cmidrule(lr){9-10}
\textbf{Model} & \textbf{Condition}
  & EER\,$(\downarrow)$ & AUC\,$(\uparrow)$
  & EER\,$(\downarrow)$ & AUC\,$(\uparrow)$
  & EER\,$(\downarrow)$ & AUC\,$(\uparrow)$
  & EER\,$(\downarrow)$ & AUC\,$(\uparrow)$\\
\midrule
\multirow{5}{*}{SentenceBERT}
  & Combined   & 0.468 & 0.543 & \deer{-0.001} & \dauc{-0.001} & \deer{ 0.003} & \dauc{-0.006} & \deer{-0.003} & \dauc{ 0.007} \\
  & Easy       & 0.222 & 0.855 & \deer{ 0.038} & \dauc{-0.033} & \deer{ 0.051} & \dauc{-0.059} & \deer{ 0.056} & \dauc{-0.066} \\
  & Hard       & 0.776 & 0.146 & \deer{-0.025} & \dauc{ 0.025} & \deer{-0.052} & \dauc{ 0.043} & \deer{-0.081} & \dauc{ 0.089} \\
  & Same Topic & 0.508 & 0.490 & \deer{-0.010} & \dauc{ 0.010} & \deer{-0.005} & \dauc{0.000} & \deer{-0.031} & \dauc{ 0.041} \\
  & Diff Topic & 0.484 & 0.515 & \deer{-0.001} & \dauc{-0.008} & \deer{ 0.018} & \dauc{-0.011} & \deer{ 0.015} & \dauc{-0.011} \\
\midrule
\multirow{5}{*}{TF-IDF}
  & Combined   & 0.482 & 0.523 & \deer{-0.004} & \dauc{ 0.003} & \deer{ 0.003} & \dauc{-0.009} & \deer{-0.013} & \dauc{ 0.021} \\
  & Easy       & 0.345 & 0.710 & \deer{ 0.036} & \dauc{-0.037} & \deer{ 0.040} & \dauc{-0.052} & \deer{ 0.022} & \dauc{-0.021} \\
  & Hard       & 0.660 & 0.280 & \deer{-0.058} & \dauc{ 0.058} & \deer{-0.040} & \dauc{ 0.041} & \deer{-0.079} & \dauc{ 0.093} \\
  & Same Topic & 0.504 & 0.489 & \deer{-0.007} & \dauc{ 0.019} & \deer{-0.002} & \dauc{-0.001} & \deer{-0.053} & \dauc{ 0.076} \\
  & Diff Topic & 0.492 & 0.508 & \deer{ 0.005} & \dauc{-0.006} & \deer{ 0.012} & \dauc{-0.013} & \deer{ 0.016} & \dauc{-0.018} \\
\midrule
\multirow{5}{*}{CISR}
  & Combined   & 0.474 & 0.535 & \deer{-0.003} & \dauc{-0.005} & \deer{ 0.002} & \dauc{-0.004} & \deer{ 0.017} & \dauc{-0.024} \\
  & Easy       & 0.308 & 0.767 & \deer{ 0.016} & \dauc{-0.030} & \deer{ 0.041} & \dauc{-0.048} & \deer{ 0.079} & \dauc{-0.108} \\
  & Hard       & 0.691 & 0.242 & \deer{-0.025} & \dauc{ 0.022} & \deer{-0.038} & \dauc{ 0.039} & \deer{-0.064} & \dauc{ 0.067} \\
  & Same Topic & 0.510 & 0.485 & \deer{-0.009} & \dauc{ 0.017} & \deer{-0.006} & \dauc{ 0.015} & \deer{ 0.004} & \dauc{-0.014} \\
  & Diff Topic & 0.480 & 0.518 & \deer{ 0.020} & \dauc{-0.016} & \deer{ 0.016} & \dauc{-0.013} & \deer{ 0.019} & \dauc{-0.019} \\
\midrule
\multirow{5}{*}{PCM}
  & Combined   & 0.468 & 0.543 & \deer{-0.005} & \dauc{ 0.006} & \deer{ 0.004} & \dauc{-0.008} & \deer{-0.005} & \dauc{ 0.005} \\
  & Easy       & 0.262 & 0.810 & \deer{ 0.014} & \dauc{-0.016} & \deer{ 0.053} & \dauc{-0.063} & \deer{ 0.047} & \dauc{-0.049} \\
  & Hard       & 0.722 & 0.202 & \deer{-0.019} & \dauc{ 0.030} & \deer{-0.038} & \dauc{ 0.048} & \deer{-0.060} & \dauc{ 0.071} \\
  & Same Topic & 0.501 & 0.498 & \deer{-0.016} & \dauc{ 0.016} & \deer{-0.002} & \dauc{-0.003} & \deer{-0.026} & \dauc{ 0.032} \\
  & Diff Topic & 0.490 & 0.519 & \deer{ 0.002} & \dauc{-0.006} & \deer{0.000} & \dauc{-0.011} & \deer{ 0.002} & \dauc{-0.016} \\
\midrule
\multirow{5}{*}{LUAR}
  & Combined   & 0.472 & 0.537 & \deer{ 0.003} & \dauc{0.000} & \deer{ 0.013} & \dauc{-0.019} & \deer{ 0.010} & \dauc{-0.021} \\
  & Easy       & 0.280 & 0.783 & \deer{ 0.033} & \dauc{-0.032} & \deer{ 0.071} & \dauc{-0.077} & \deer{ 0.096} & \dauc{-0.118} \\
  & Hard       & 0.723 & 0.217 & \deer{-0.028} & \dauc{ 0.036} & \deer{-0.058} & \dauc{ 0.052} & \deer{-0.098} & \dauc{ 0.099} \\
  & Same Topic & 0.502 & 0.492 & \deer{ 0.006} & \dauc{-0.003} & \deer{ 0.015} & \dauc{-0.014} & \deer{ 0.021} & \dauc{-0.023} \\
  & Diff Topic & 0.487 & 0.521 & \deer{ 0.001} & \dauc{0.000} & \deer{ 0.017} & \dauc{-0.024} & \deer{ 0.004} & \dauc{-0.015} \\
\midrule
\multirow{5}{*}{GPT-5-mini}
  & Combined   & \textbf{0.379} & \textbf{0.664} & \deer{ 0.036} & \dauc{-0.042} & \deer{ 0.054} & \dauc{-0.054} & \deer{ 0.060} & \dauc{-0.077} \\
  & Easy       & 0.222 & 0.832 & \deer{ 0.041} & \dauc{-0.026} & \deer{ 0.088} & \dauc{-0.072} & \deer{ 0.108} & \dauc{-0.113} \\
  & Hard       & \textbf{0.589} & \textbf{0.431} & \deer{ 0.024} & \dauc{-0.056} & \deer{ 0.011} & \dauc{-0.028} & \deer{0.000} & \dauc{-0.030} \\
  & Same Topic & 0.437 & 0.608 & \deer{-0.011} & \dauc{-0.007} & \deer{ 0.021} & \dauc{-0.036} & \deer{ 0.036} & \dauc{-0.061} \\
  & Diff Topic & \textbf{0.377} & \textbf{0.669} & \deer{ 0.060} & \dauc{-0.071} & \deer{ 0.062} & \dauc{-0.071} & \deer{ 0.072} & \dauc{-0.086} \\
\midrule
\multirow{5}{*}{Llama-3.3-70B}
  & Combined   & 0.403 & 0.632 & \deer{ 0.012} & \dauc{-0.016} & \deer{ 0.034} & \dauc{-0.030} & \deer{ 0.048} & \dauc{-0.059} \\
  & Easy       & \textbf{0.192} & \textbf{0.868} & \deer{ 0.030} & \dauc{-0.021} & \deer{ 0.111} & \dauc{-0.103} & \deer{ 0.122} & \dauc{-0.115} \\
  & Hard       & 0.668 & 0.313 & \deer{-0.005} & \dauc{-0.014} & \deer{-0.056} & \dauc{ 0.069} & \deer{-0.039} & \dauc{ 0.017} \\
  & Same Topic & \textbf{0.422} & \textbf{0.611} & \deer{ 0.015} & \dauc{-0.022} & \deer{ 0.031} & \dauc{-0.019} & \deer{ 0.043} & \dauc{-0.055} \\
  & Diff Topic & 0.420 & 0.612 & \deer{ 0.018} & \dauc{-0.023} & \deer{ 0.024} & \dauc{-0.040} & \deer{ 0.048} & \dauc{-0.069} \\
\midrule
\multirow{5}{*}{Qwen3.5-9B}
  & Combined   & 0.478 & 0.528 & \deer{-0.013} & \dauc{ 0.020} & \deer{ 0.008} & \dauc{-0.006} & \deer{ 0.013} & \dauc{-0.016} \\
  & Easy       & 0.300 & 0.744 & \deer{-0.008} & \dauc{ 0.018} & \deer{ 0.122} & \dauc{-0.128} & \deer{ 0.110} & \dauc{-0.121} \\
  & Hard       & 0.699 & 0.257 & \deer{-0.009} & \dauc{ 0.008} & \deer{-0.121} & \dauc{ 0.146} & \deer{-0.106} & \dauc{ 0.111} \\
  & Same Topic & 0.498 & 0.499 & \deer{-0.016} & \dauc{ 0.020} & \deer{-0.015} & \dauc{ 0.023} & \deer{-0.003} & \dauc{ 0.001} \\
  & Diff Topic & 0.499 & 0.504 & \deer{-0.016} & \dauc{ 0.018} & \deer{ 0.005} & \dauc{-0.004} & \deer{0.000} & \dauc{-0.012} \\
\bottomrule
\end{tabular}
\caption{Biological sex verification results on NaturalVoices for all models across trial conditions and ablation settings.
Main results use probe\,$\approx$\,200 words and $L=20$;
$\Delta$ columns show the change from the main setting.
Best EER and AUC per condition (main setting) in \textbf{bold}.}
\label{tab:sex_full}
\end{table*}

\begin{table*}[h]
\centering
\small
\setlength{\tabcolsep}{3.5pt}
\begin{tabular}{ll rr rr rr rr}
\toprule
& & \multicolumn{2}{c}{\textbf{Main - 200w / }$L$\textbf{=20}}
  & \multicolumn{2}{c}{$\Delta$ \textbf{200w / }$L$\textbf{=5}}
  & \multicolumn{2}{c}{$\Delta$ \textbf{50w / }$L$\textbf{=20}}
  & \multicolumn{2}{c}{$\Delta$ \textbf{50w / }$L$\textbf{=5}} \\
\cmidrule(lr){3-4}\cmidrule(lr){5-6}\cmidrule(lr){7-8}\cmidrule(lr){9-10}
\textbf{Model} & \textbf{Condition}
  & EER\,$(\downarrow)$ & AUC\,$(\uparrow)$
  & EER\,$(\downarrow)$ & AUC\,$(\uparrow)$
  & EER\,$(\downarrow)$ & AUC\,$(\uparrow)$
  & EER\,$(\downarrow)$ & AUC\,$(\uparrow)$\\
\midrule
\multirow{5}{*}{SentenceBERT}
  & Combined   & 0.447 & 0.576 & \deer{ 0.037} & \dauc{-0.052} & \deer{ 0.023} & \dauc{-0.039} & \deer{ 0.026} & \dauc{-0.043} \\
  & Easy       & \textbf{0.198} & \textbf{0.873} & \deer{ 0.051} & \dauc{-0.045} & \deer{ 0.063} & \dauc{-0.060} & \deer{ 0.091} & \dauc{-0.092} \\
  & Hard       & 0.746 & 0.173 & \deer{ 0.002} & \dauc{0.000} & \deer{-0.033} & \dauc{ 0.037} & \deer{-0.053} & \dauc{ 0.065} \\
  & Same Topic & 0.461 & 0.545 & \deer{ 0.030} & \dauc{-0.049} & \deer{ 0.033} & \dauc{-0.038} & \deer{ 0.052} & \dauc{-0.051} \\
  & Diff Topic & 0.469 & 0.548 & \deer{ 0.018} & \dauc{-0.037} & \deer{ 0.022} & \dauc{-0.033} & \deer{ 0.014} & \dauc{-0.028} \\
\midrule
\multirow{5}{*}{TF-IDF}
  & Combined   & 0.478 & 0.523 & \deer{ 0.026} & \dauc{-0.028} & \deer{ 0.026} & \dauc{-0.026} & \deer{ 0.025} & \dauc{-0.029} \\
  & Easy       & 0.343 & 0.706 & \deer{ 0.051} & \dauc{-0.049} & \deer{ 0.053} & \dauc{-0.052} & \deer{ 0.081} & \dauc{-0.092} \\
  & Hard       & \textbf{0.661} & \textbf{0.277} & \deer{-0.032} & \dauc{ 0.034} & \deer{-0.019} & \dauc{ 0.036} & \deer{-0.056} & \dauc{ 0.076} \\
  & Same Topic & 0.517 & 0.478 & \deer{ 0.001} & \dauc{-0.003} & \deer{ 0.004} & \dauc{0.000} & \deer{ 0.005} & \dauc{-0.006} \\
  & Diff Topic & 0.488 & 0.517 & \deer{ 0.022} & \dauc{-0.026} & \deer{ 0.018} & \dauc{-0.031} & \deer{ 0.018} & \dauc{-0.028} \\
\midrule
\multirow{5}{*}{CISR}
  & Combined   & 0.460 & 0.556 & \deer{ 0.022} & \dauc{-0.030} & \deer{ 0.009} & \dauc{-0.014} & \deer{ 0.024} & \dauc{-0.026} \\
  & Easy       & 0.294 & 0.781 & \deer{ 0.025} & \dauc{-0.035} & \deer{ 0.034} & \dauc{-0.041} & \deer{ 0.061} & \dauc{-0.073} \\
  & Hard       & 0.672 & 0.266 & \deer{-0.012} & \dauc{ 0.009} & \deer{-0.043} & \dauc{ 0.046} & \deer{-0.047} & \dauc{ 0.057} \\
  & Same Topic & 0.457 & 0.552 & \deer{ 0.041} & \dauc{-0.039} & \deer{ 0.011} & \dauc{-0.015} & \deer{ 0.013} & \dauc{-0.020} \\
  & Diff Topic & 0.476 & 0.521 & \deer{ 0.013} & \dauc{-0.005} & \deer{ 0.002} & \dauc{0.000} & \deer{ 0.017} & \dauc{-0.016} \\
\midrule
\multirow{5}{*}{PCM}
  & Combined   & 0.443 & \textbf{0.584} & \deer{ 0.043} & \dauc{-0.066} & \deer{ 0.025} & \dauc{-0.049} & \deer{ 0.032} & \dauc{-0.055} \\
  & Easy       & 0.246 & 0.835 & \deer{ 0.039} & \dauc{-0.047} & \deer{ 0.055} & \dauc{-0.072} & \deer{ 0.070} & \dauc{-0.090} \\
  & Hard       & 0.696 & 0.240 & \deer{ 0.017} & \dauc{-0.020} & \deer{-0.029} & \dauc{ 0.023} & \deer{-0.033} & \dauc{ 0.031} \\
  & Same Topic & \textbf{0.440} & \textbf{0.576} & \deer{ 0.056} & \dauc{-0.070} & \deer{ 0.047} & \dauc{-0.063} & \deer{ 0.045} & \dauc{-0.071} \\
  & Diff Topic & \textbf{0.447} & 0.556 & \deer{ 0.056} & \dauc{-0.059} & \deer{ 0.041} & \dauc{-0.039} & \deer{ 0.045} & \dauc{-0.042} \\
\midrule
\multirow{5}{*}{LUAR}
  & Combined   & 0.462 & 0.553 & \deer{ 0.023} & \dauc{-0.034} & \deer{ 0.014} & \dauc{-0.032} & \deer{ 0.033} & \dauc{-0.039} \\
  & Easy       & 0.269 & 0.794 & \deer{ 0.037} & \dauc{-0.041} & \deer{ 0.065} & \dauc{-0.076} & \deer{ 0.109} & \dauc{-0.123} \\
  & Hard       & 0.704 & 0.236 & \deer{-0.012} & \dauc{ 0.011} & \deer{-0.060} & \dauc{ 0.052} & \deer{-0.074} & \dauc{ 0.096} \\
  & Same Topic & 0.467 & 0.531 & \deer{ 0.033} & \dauc{-0.034} & \deer{ 0.034} & \dauc{-0.028} & \deer{ 0.030} & \dauc{-0.030} \\
  & Diff Topic & 0.475 & 0.525 & \deer{ 0.010} & \dauc{-0.011} & \deer{ 0.013} & \dauc{-0.020} & \deer{ 0.025} & \dauc{-0.026} \\
\midrule
\multirow{5}{*}{GPT-5-mini}
  & Combined   & \textbf{0.442} & 0.572 & \deer{ 0.026} & \dauc{-0.023} & \deer{ 0.030} & \dauc{-0.032} & \deer{ 0.026} & \dauc{-0.026} \\
  & Easy       & 0.227 & 0.843 & \deer{ 0.021} & \dauc{-0.017} & \deer{ 0.063} & \dauc{-0.069} & \deer{ 0.062} & \dauc{-0.065} \\
  & Hard       & 0.728 & 0.210 & \deer{0.000} & \dauc{ 0.002} & \deer{-0.045} & \dauc{ 0.038} & \deer{-0.047} & \dauc{ 0.046} \\
  & Same Topic & 0.496 & 0.510 & \deer{ 0.011} & \dauc{-0.025} & \deer{-0.003} & \dauc{-0.011} & \deer{-0.006} & \dauc{-0.004} \\
  & Diff Topic & 0.459 & \textbf{0.557} & \deer{-0.009} & \dauc{ 0.008} & \deer{ 0.008} & \dauc{-0.020} & \deer{ 0.004} & \dauc{-0.014} \\
\midrule
\multirow{5}{*}{Llama-3.3-70B}
  & Combined   & 0.546 & 0.433 & \deer{-0.070} & \dauc{ 0.091} & \deer{-0.075} & \dauc{ 0.098} & \deer{-0.065} & \dauc{ 0.090} \\
  & Easy       & 0.217 & 0.856 & \deer{ 0.002} & \dauc{-0.004} & \deer{ 0.063} & \dauc{-0.049} & \deer{ 0.048} & \dauc{-0.053} \\
  & Hard       & 0.764 & 0.143 & \deer{ 0.011} & \dauc{-0.008} & \deer{-0.044} & \dauc{ 0.048} & \deer{-0.017} & \dauc{ 0.042} \\
  & Same Topic & 0.542 & 0.474 & \deer{-0.004} & \dauc{-0.007} & \deer{-0.012} & \dauc{ 0.001} & \deer{-0.011} & \dauc{-0.009} \\
  & Diff Topic & 0.481 & 0.535 & \deer{-0.006} & \dauc{-0.005} & \deer{-0.011} & \dauc{-0.005} & \deer{-0.001} & \dauc{-0.016} \\
\midrule
\multirow{5}{*}{Qwen3.5-9B}
  & Combined   & 0.470 & 0.537 & \deer{ 0.020} & \dauc{-0.036} & \deer{ 0.029} & \dauc{-0.042} & \deer{ 0.026} & \dauc{-0.036} \\
  & Easy       & 0.253 & 0.804 & \deer{ 0.009} & \dauc{-0.019} & \deer{ 0.119} & \dauc{-0.137} & \deer{ 0.124} & \dauc{-0.132} \\
  & Hard       & 0.755 & 0.195 & \deer{ 0.004} & \dauc{-0.015} & \deer{-0.093} & \dauc{ 0.093} & \deer{-0.104} & \dauc{ 0.103} \\
  & Same Topic & 0.516 & 0.471 & \deer{ 0.032} & \dauc{-0.021} & \deer{ 0.026} & \dauc{-0.024} & \deer{ 0.012} & \dauc{-0.009} \\
  & Diff Topic & 0.486 & 0.518 & \deer{ 0.017} & \dauc{-0.018} & \deer{ 0.019} & \dauc{-0.021} & \deer{ 0.016} & \dauc{-0.021} \\
\bottomrule
\end{tabular}
\caption{Age-group verification results on NaturalVoices for all models across trial conditions and ablation settings.
Main results use probe\,$\approx$\,200 words and $L=20$;
$\Delta$ columns show the change from the main setting.
Best EER and AUC per condition (main setting) in \textbf{bold}.}
\label{tab:age_full}
\end{table*}

\begin{table*}[h]
\centering
\small
\setlength{\tabcolsep}{3.5pt}
\begin{tabular}{ll rr rr rr rr}
\toprule
& & \multicolumn{2}{c}{\textbf{Main - 200w / }$L$\textbf{=20}}
  & \multicolumn{2}{c}{$\Delta$ \textbf{200w / }$L$\textbf{=5}}
  & \multicolumn{2}{c}{$\Delta$ \textbf{50w / }$L$\textbf{=20}}
  & \multicolumn{2}{c}{$\Delta$ \textbf{50w / }$L$\textbf{=5}} \\
\cmidrule(lr){3-4}\cmidrule(lr){5-6}\cmidrule(lr){7-8}\cmidrule(lr){9-10}
\textbf{Model} & \textbf{Condition}
  & EER\,$(\downarrow)$ & AUC\,$(\uparrow)$
  & EER\,$(\downarrow)$ & AUC\,$(\uparrow)$
  & EER\,$(\downarrow)$ & AUC\,$(\uparrow)$
  & EER\,$(\downarrow)$ & AUC\,$(\uparrow)$\\
\midrule
\multirow{5}{*}{SentenceBERT}
  & Combined   & 0.441 & 0.578 & \deer{-0.002} & \dauc{ 0.010} & \deer{ 0.004} & \dauc{-0.001} & \deer{ 0.010} & \dauc{-0.014} \\
  & Easy       & \textbf{0.212} & \textbf{0.869} & \deer{ 0.033} & \dauc{-0.027} & \deer{ 0.045} & \dauc{-0.053} & \deer{ 0.074} & \dauc{-0.086} \\
  & Hard       & 0.752 & 0.171 & \deer{-0.039} & \dauc{ 0.045} & \deer{-0.063} & \dauc{ 0.060} & \deer{-0.076} & \dauc{ 0.083} \\
  & Same Topic & 0.483 & 0.534 & \deer{-0.023} & \dauc{ 0.018} & \deer{-0.010} & \dauc{ 0.003} & \deer{-0.004} & \dauc{-0.012} \\
  & Diff Topic & 0.494 & 0.510 & \deer{-0.021} & \dauc{ 0.021} & \deer{-0.009} & \dauc{ 0.011} & \deer{-0.005} & \dauc{ 0.003} \\
\midrule
\multirow{5}{*}{TF-IDF}
  & Combined   & 0.465 & 0.545 & \deer{ 0.013} & \dauc{-0.015} & \deer{ 0.020} & \dauc{-0.018} & \deer{ 0.033} & \dauc{-0.031} \\
  & Easy       & 0.335 & 0.715 & \deer{ 0.055} & \dauc{-0.053} & \deer{ 0.058} & \dauc{-0.059} & \deer{ 0.099} & \dauc{-0.102} \\
  & Hard       & 0.620 & 0.328 & \deer{-0.050} & \dauc{ 0.048} & \deer{-0.034} & \dauc{ 0.051} & \deer{-0.059} & \dauc{ 0.075} \\
  & Same Topic & 0.471 & 0.539 & \deer{-0.004} & \dauc{ 0.003} & \deer{-0.013} & \dauc{ 0.005} & \deer{ 0.006} & \dauc{-0.008} \\
  & Diff Topic & 0.499 & 0.494 & \deer{ 0.020} & \dauc{-0.015} & \deer{ 0.017} & \dauc{-0.021} & \deer{ 0.032} & \dauc{-0.023} \\
\midrule
\multirow{5}{*}{CISR}
  & Combined   & 0.446 & 0.572 & \deer{ 0.025} & \dauc{-0.025} & \deer{ 0.003} & \dauc{-0.006} & \deer{ 0.013} & \dauc{-0.015} \\
  & Easy       & 0.289 & 0.786 & \deer{ 0.040} & \dauc{-0.049} & \deer{ 0.032} & \dauc{-0.045} & \deer{ 0.056} & \dauc{-0.068} \\
  & Hard       & 0.663 & 0.270 & \deer{-0.002} & \dauc{ 0.015} & \deer{-0.056} & \dauc{ 0.054} & \deer{-0.050} & \dauc{ 0.059} \\
  & Same Topic & 0.474 & 0.546 & \deer{ 0.004} & \dauc{-0.010} & \deer{ 0.002} & \dauc{0.000} & \deer{ 0.014} & \dauc{-0.018} \\
  & Diff Topic & 0.486 & 0.520 & \deer{ 0.020} & \dauc{-0.027} & \deer{-0.005} & \dauc{-0.001} & \deer{0.000} & \dauc{-0.002} \\
\midrule
\multirow{5}{*}{PCM}
  & Combined   & 0.438 & 0.594 & \deer{-0.005} & \dauc{-0.009} & \deer{ 0.004} & \dauc{-0.011} & \deer{ 0.001} & \dauc{-0.012} \\
  & Easy       & 0.239 & 0.839 & \deer{ 0.025} & \dauc{-0.037} & \deer{ 0.054} & \dauc{-0.066} & \deer{ 0.060} & \dauc{-0.078} \\
  & Hard       & 0.700 & 0.243 & \deer{-0.047} & \dauc{ 0.028} & \deer{-0.066} & \dauc{ 0.067} & \deer{-0.069} & \dauc{ 0.082} \\
  & Same Topic & 0.432 & 0.573 & \deer{ 0.018} & \dauc{-0.002} & \deer{ 0.012} & \dauc{-0.007} & \deer{ 0.026} & \dauc{-0.006} \\
  & Diff Topic & 0.481 & 0.534 & \deer{ 0.002} & \dauc{-0.005} & \deer{-0.006} & \dauc{0.000} & \deer{-0.009} & \dauc{0.000} \\
\midrule
\multirow{5}{*}{LUAR}
  & Combined   & 0.446 & 0.576 & \deer{ 0.007} & \dauc{-0.010} & \deer{ 0.018} & \dauc{-0.023} & \deer{ 0.026} & \dauc{-0.037} \\
  & Easy       & 0.265 & 0.793 & \deer{ 0.037} & \dauc{-0.034} & \deer{ 0.074} & \dauc{-0.064} & \deer{ 0.108} & \dauc{-0.113} \\
  & Hard       & 0.676 & 0.275 & \deer{-0.037} & \dauc{ 0.025} & \deer{-0.041} & \dauc{ 0.032} & \deer{-0.055} & \dauc{ 0.059} \\
  & Same Topic & 0.479 & 0.540 & \deer{-0.017} & \dauc{ 0.011} & \deer{ 0.027} & \dauc{-0.035} & \deer{ 0.032} & \dauc{-0.045} \\
  & Diff Topic & 0.481 & 0.523 & \deer{ 0.013} & \dauc{-0.011} & \deer{ 0.012} & \dauc{-0.010} & \deer{ 0.016} & \dauc{-0.012} \\
\midrule
\multirow{5}{*}{GPT-5-mini}
  & Combined   & \textbf{0.279} & \textbf{0.763} & \deer{ 0.022} & \dauc{-0.024} & \deer{ 0.039} & \dauc{-0.036} & \deer{ 0.059} & \dauc{-0.057} \\
  & Easy       & 0.227 & 0.855 & \deer{ 0.008} & \dauc{-0.028} & \deer{ 0.029} & \dauc{-0.051} & \deer{ 0.067} & \dauc{-0.082} \\
  & Hard       & \textbf{0.434} & \textbf{0.584} & \deer{-0.025} & \dauc{ 0.001} & \deer{-0.007} & \dauc{ 0.008} & \deer{-0.020} & \dauc{ 0.005} \\
  & Same Topic & 0.350 & \textbf{0.701} & \deer{ 0.017} & \dauc{-0.005} & \deer{ 0.022} & \dauc{-0.013} & \deer{ 0.022} & \dauc{-0.037} \\
  & Diff Topic & \textbf{0.268} & \textbf{0.771} & \deer{ 0.034} & \dauc{-0.033} & \deer{ 0.044} & \dauc{-0.044} & \deer{ 0.069} & \dauc{-0.064} \\
\midrule
\multirow{5}{*}{Llama-3.3-70B}
  & Combined   & 0.315 & 0.720 & \deer{ 0.023} & \dauc{-0.015} & \deer{ 0.053} & \dauc{-0.035} & \deer{ 0.064} & \dauc{-0.045} \\
  & Easy       & 0.241 & 0.818 & \deer{ 0.019} & \dauc{-0.004} & \deer{ 0.060} & \dauc{-0.060} & \deer{ 0.070} & \dauc{-0.067} \\
  & Hard       & 0.460 & 0.554 & \deer{-0.013} & \dauc{-0.018} & \deer{ 0.002} & \dauc{ 0.016} & \deer{ 0.011} & \dauc{ 0.005} \\
  & Same Topic & \textbf{0.332} & 0.699 & \deer{ 0.009} & \dauc{ 0.005} & \deer{ 0.043} & \dauc{-0.027} & \deer{ 0.045} & \dauc{-0.030} \\
  & Diff Topic & 0.320 & 0.707 & \deer{ 0.035} & \dauc{-0.029} & \deer{ 0.032} & \dauc{-0.036} & \deer{ 0.053} & \dauc{-0.051} \\
\midrule
\multirow{5}{*}{Qwen3.5-9B}
  & Combined   & 0.441 & 0.566 & \deer{-0.003} & \dauc{0.000} & \deer{ 0.016} & \dauc{-0.014} & \deer{ 0.032} & \dauc{-0.026} \\
  & Easy       & 0.275 & 0.773 & \deer{ 0.019} & \dauc{-0.033} & \deer{ 0.105} & \dauc{-0.110} & \deer{ 0.124} & \dauc{-0.134} \\
  & Hard       & 0.695 & 0.268 & \deer{-0.048} & \dauc{ 0.044} & \deer{-0.117} & \dauc{ 0.119} & \deer{-0.121} & \dauc{ 0.134} \\
  & Same Topic & 0.495 & 0.518 & \deer{-0.006} & \dauc{ 0.012} & \deer{-0.015} & \dauc{ 0.009} & \deer{-0.014} & \dauc{ 0.010} \\
  & Diff Topic & 0.480 & 0.524 & \deer{-0.015} & \dauc{ 0.006} & \deer{0.000} & \dauc{ 0.005} & \deer{ 0.013} & \dauc{-0.010} \\
\bottomrule
\end{tabular}
\caption{Accent verification results on NaturalVoices for all models across trial conditions and ablation settings.
Main results use probe\,$\approx$\,200 words and $L=20$;
$\Delta$ columns show the change from the main setting.
Best EER and AUC per condition (main setting) in \textbf{bold}.}
\label{tab:accent_full}
\end{table*}

\end{document}